\documentclass[prodmode,acmtissec]{acmsmall}
\usepackage{amssymb,graphicx,bezier,amsfonts,enumerate,latexsym,amsmath}
\usepackage{epstopdf}
\usepackage{listings}
\usepackage{floatflt}
\usepackage{subfigure}
\usepackage{url}
\usepackage{program}
\usepackage{multirow}
\usepackage{epsfig}
\usepackage[ruled]{algorithm2e}
\renewcommand{\algorithmcfname}{ALGORITHM}
\SetAlFnt{\small}
\SetAlCapFnt{\small}
\SetAlCapNameFnt{\small}
\SetAlCapHSkip{0pt}
\usepackage{setspace}
\IncMargin{-\parindent}
\usepackage{csquotes}
\begin{document}
	\markboth{R. Kumar et al.}{Authenticating users through their arm movement patterns}
	\title{Authenticating users through their arm movement patterns}
	\author{Rajesh Kumar
    \affil{Syracuse University, NY, USA}
    Vir V Phoha
    \affil{Syracuse University, NY, USA}
    Rahul Raina
    \affil{\affil{Georgia Institute of Technology, GA, USA}}
}
\begin{abstract}
In this paper, we propose four continuous authentication designs by using the characteristics of arm movements while individuals walk. The first design uses acceleration of arms captured by a smartwatch's accelerometer sensor, the second design uses the rotation of arms captured by a smartwatch's gyroscope sensor, third uses the fusion of both acceleration and rotation at the feature-level and fourth uses the fusion at score-level. Each of these designs is implemented by using four classifiers, namely, k nearest neighbors (k-NN) with Euclidean distance, Logistic Regression, Multilayer Perceptrons, and Random Forest resulting in a total of sixteen authentication mechanisms. These authentication mechanisms are tested under three different environments, namely an intra-session, inter-session on a dataset of 40 users and an inter-phase on a dataset of 12 users. The sessions of data collection were separated by at least ten minutes, whereas the phases of data collection were separated by at least three months. Under the intra-session environment, all of the twelve authentication mechanisms achieve a mean dynamic false accept rate (DFAR) of 0\% and dynamic false reject rate (DFRR) of 0\%. For the inter-session environment, feature level fusion-based design with classifier k-NN achieves the best error rates that are a mean DFAR of 2.2\% and DFRR of 4.2\%. The DFAR and DFRR increased from 5.68\% and 4.23\% to 15.03\% and 14.62\% respectively when feature level fusion-based design with classifier k-NN was tested under the inter-phase environment on a dataset of 12 users.
\end{abstract}
\maketitle
\section{Introduction}With the emergence of the Internet of Things (IoT), cyber physical objects such as smart vehicles, smart buildings, smart devices and other things are to be sensed and controlled remotely across existing network infrastructure. The IoT provides an interface for direct integration of the physical world into computer-based systems. When augmented with sensors and actuators, the IoT also improves the efficiency and accuracy of the objects connected to it \cite{WikipediaIoT}. However, these benefits come with a high risk of security and privacy. Individuals with malicious intent can get unauthorized access to IoT devices and may create havoc. Therefore, developing authentication mechanisms for users who are authorized to access these devices is essential. Developing a foolproof authentication mechanism is extremely challenging due to the following. Firstly, users of these devices strive for secure and less time consuming authentication mechanisms to accommodate the need of their ever quickening lives. Secondly, the most prominent authentication mechanism based on PINs and passwords, are under question as they can be stolen by utilizing various side channels \cite{WristSnoop}\cite{ShuklaAttack}. Finally, despite being faster and easier to use, the physiological biometric, e.g. fingerprint, face, and iris-based authentication mechanisms suffer from two weaknesses: i) they provide only entry (or one) point authentication which provides a window for unauthorized access;, and, ii) they are susceptible to spoof attacks \cite{FingerPrintScanner}. For instance, if the owners keep their device unlocked and unattended or if they are sleeping, intoxicated or unconscious, their fingerprint can be obtained easily to unlock the device.\\
Alternatively, researchers have been exploring the possibility of authenticating users of these devices continuously based on their behaviometrics\footnote{Any human behavioral characteristic that satisfies the following requirements: universality, distinctiveness, permanence, collectability, performance, acceptability, and circumvention; for example, typing, swiping, and walking \cite{JainBiometrics}.}, especially swiping, typing, and gait (walking patterns) \cite{PhoneSwiping1}\cite{PhoneSwiping2}\cite{PhoneTyping2}\cite{PhoneGait2}. Of these, gait captured by a smartphone accelerometer has shown promise and is seen to be a viable means for authenticating users on smartphones \cite{AuthenticationForPhones} as it achieves a significantly high accuracy. However, most studies of gait biometrics assume that the phone is placed at a fixed location e.g. pocket, hand, or waist. This is a strong assumption because it ignores the variations introduced (due to changes in the placement of the phone) in the walking pattern as captured by accelerometer. For instance, Primo et al. \cite{ABENA} demonstrated that variations in acceleration caused by changing the position of the phone from pocket to hand affects the authentication accuracy markedly \cite{ABENA}. The researchers proposed a multi-stage authentication framework in which the system creates different templates for different locations of the phone. However, to implement the idea in a real scenario, the system would need to identify the exact location of the phone on the owner's body. Any error in prediction of the phone location could result in an incorrect authentication decision. Unfortunately, no proper framework exists so far, that can locate the exact position of the phone automatically.\\
In contrast to smartphones, smartwatches are always worn on the wrist by their users, provide a more consistent source for capturing arm movements while walking and may provide a potential basis for authenticating users. If arm movements can be used as a behaviometric, not only can it be used to authenticate users to access smartwatches, but also smartphones or any other device paired with a smartwatch \cite{AuthenticatePhoneWithWatch}. In addition, it is crucial to study security through smartwatches because they are seen as the potential replacement for smartphones in the near future \cite{ReplaceSmartphones}. Samsung, Apple and Microsoft have already begun to incorporate novel sensors and features into their smartwatches. For example, Samsung is planning to add biosensors to their Gear S that would analyze wrist rotations and heartbeat signals to authenticate users \cite{SmartwatchPayments}.\\
Thus, this paper examines whether acceleration and rotation generated by an individual's arm movements while walking, as captured through an accelerometer and a gyroscope built into a smartwatch, can be used to authenticate users. We also explore whether the fusion of these two (i.e. acceleration and rotation) enhance authentication performance.\\
Our contributions are as follows:
\begin{itemize}
  \item Following our university's institutional review board (IRB) guidelines, we built a dataset of arm movements of individuals walking naturally. The data collection experiment was carried out in two different phases. A total of 40 individuals participated in the first phase. Twelve of them followed up in the second phase which was carried out after three months. We plan to share our data, application and supporting code publicly in order to facilitate fellow researchers to reproduce the results for further investigations or comparative studies. For details, see Section \ref{sec:DataCollection}.
  \item We extracted a total of 32 features from accelerometer readings and a total of 44 features from gyroscope readings. By using two feature evaluation methods, namely, the information gain based feature ranking (IGFR) and the correlation based feature subset selection (CFSS) methods, we evaluate the importance of these features. The IGFR method helped in ranking the features according to their discriminability, whereas the CFSS method helped in selecting the best subset of features for classification. The feature evaluation helped us discard more than 25\% of the features. The effect of feature selection is demonstrated by comparing the performance of the classifiers with and without feature selection. For details, see Section \ref{sec:FeatureAnalysis}.
  \item An empirical analysis of two important parameters i.e. the window size ($W_{size}$) used for feature extraction and the amount of overlap or sliding interval ($S_{interval}$) among the consecutive windows is performed. These two parameters are critical for a continuous authentication mechanism as the former decides the time taken to give the first authentication decision, whereas the latter determines the time taken to give subsequent decisions. We propose optimal settings for these two parameters are suggested. For more details, see Section \ref{subsubsec:LengthOfWindow}.
  \item We propose three different continuous authentication designs based on the characteristics of arm movements of individuals: first, by using only acceleration, second, by utilizing only rotation, and third by fusing these two at the feature level. Each of these designs are implemented by four classification algorithms, namely, k nearest neighbors (kNN) with Euclidean distance, logistic regression, multilayer perceptrons, and random forest. Consequently, we propose a total of twelve authentication mechanisms. All of these mechanisms are tested in three different environments, namely, intra-session, inter-session and inter-phase. The performance of all of them are presented in terms of DFAR, DFRR and dynamic accuracy. For more details, see Section \ref{sec:ExperimentalDesign}.
\end{itemize}
The rest of this paper is organized as follows: the related literature is discussed in Section \ref{sec:RelatedWork}; continuous authentication and threat model is discussed in Section \ref{sec:ContinuousAuthentication}; data collection and feature analysis are presented in Sections \ref{sec:DataCollection} and \ref{sec:FeatureAnalysis}; the experimental design and performance evaluation methods are described in Section \ref{sec:ExperimentalDesign}; and, the conclusions and ideas for future work are given in Section \ref{sec:Conclusion}.
\section{Related Work}
\label{sec:RelatedWork}
To the best of our knowledge, the most closely related work came out by Johnston et al. \cite{FordhamSmartWatch} while our work was in progress. In their paper, Johnston et al. also propose similar authentication mechanisms as ours. However there are several factors that distinguish our work apart from theirs.\\ First, they only test their systems in an intra-session environment. The intra-session environment does not represent a realistic scenario as we can not carry out the enrollment and verification in the same session. The intra session environment may favor the classification process and result in high accuracy. We also carried out performance evaluation of the intra-session environment, which always resulted in perfect classification accuracies (error rate of 0\%). We assume a more realistic scenario in which the enrollment is carried out by using the data collected in one session and verification by using the other. We refer to this as the inter-session environment. Additionally, we test our system in a far more realistic scenario in which the verification is carried out after three months of the enrollment. Johnston et al. mention that when they tried to evaluate the performance in the intra-session or inter-phase environments, the error rates were very high. Contrary to their study, our design of authentication not only performs well in the inter-session and but also achieves state-of-the-art classification error rates in the inter-phase environment.\\
Second, the feature set used by Johnston et al. contains mostly statistical features that have mainly been used for smartphone accelerometers in the past \cite{PhoneGait1}. The same set of features were extracted from gyroscope readings as well. Our feature set  consists of features from both a time and frequency domain. We also define sixteen novel features dedicated to the rotation (see Table \ref{ListOfTotalFeatures}). Furthermore, Johnston et al. did not carry out any feature analysis in order to find out the strength of the features. However, we carry out an extensive feature analysis by using two prominent methods IGFR and CFSS (see Section \ref{sec:FeatureExtraction}). Through this process, we were able to discard more than 25\% of the total features which, not only resulted in improved classification accuracy but also reduced classification time.\\
Third, we use two different classification algorithms, namely, k-NN and Logistic Regression. Interestingly, both of these outperform the other common ones (i.e., multilayer perceptrons and random forest) in terms of classification accuracy (see Figure \ref{PerformanceComparision} and Table \ref{InterPhaseINterSession}).\\
Finally, and more importantly, we carry out a fusion of acceleration and rotation at the feature level, a consideration that is stated in their future work. We also provide an empirical evidence that our system is scalable to a large population of users which is an important aspect that has not been studied by Johnston et al.\\
In addition, researchers have studied wrist motion while making pre-defined gestures such as geometrical shape (circle, triangle etc.) or any alphabet or word for authentication purposes. The wrist motions in these studies are captured by either wrist worn sensors or an accelerometer and gyroscope built into smartwatches. For example, Yang et al. \cite{HMMSmartWatch} studied four different types of gestures, namely, circle, up, down, and rotation that were collected from 30 users for authenticating them. The authors apply two different methods, namely, histogram and dynamic time warping (DTW), and report the distribution of equal error rate (EER) across the users. Above 70\% of the users hit an EER of $\sim$ 5\%, whereas the remaining users' EER lies between 5 to 25\%. Their proposed authentication system \cite{HMMSmartWatch} is an \textit{entry point} authentication system, which means it does not monitor the device continuously. Whereas, in our paper, we propose continuous authentication systems based on arm movements while walking.\\
Guiry et al. \cite{FusionOfMultipleSensorsAccGyro}, used smartwatches and smartphones to address the activity recognition problem. They conclude that although acceleration contributes significantly in human activity recognition, rotation helps in improving the overall performance. Additionally, Lorenzo et al. \cite{SmartWatchGestureRecognition} used smartwatch to develop an assistance mechanism for visually challenged people during their daily activities. In their paper, they develop a gesture recognition system based on the combination of signals received from the smartwatch and smartphone together. Moreover, a variety of applications of smartwatches are being proposed by researchers. For example, smartwatch-based payment \cite{SmartwatchPayments}, smartwatch based authentication of users to access phones or other paired devices \cite{AuthenticatePhoneWithWatch} and fall detection in elderly people \cite{ETelemed}.\\
Although our proposed authentication framework relies solely upon arm movements \textit{while walking}, we believe that it can be easily extended to arm movements while running, while cycling, stair ascents and stair descents or while typing by designing a multi-stage authentication framework. The multi-stage authentication framework will use different template for different activities similar to the one proposed by Primo et al. \cite{ABENA}. However, the framework will need to identify the activities on the fly. The error in activity classification will only reduce the authentication accuracy.
\section{Continuous Authentication}
\label{sec:ContinuousAuthentication}
Continuous authentication (CA) is a process of repeatedly verifying the identity of individuals (who are authorized to use a device or a system) . The verification is carried out at predefined or random intervals, or after the occurrence of specific events. The predefined interval usually depends upon the availability, quality and quantity of data to be used as input for CA. The CA framework mainly consists of four components: enrollment, repeated verification, re-login, and template update \cite{ContinuousAuthentication}. During the enrollment, a template (profile) for the genuine user is created. Sometimes multiple templates are also created to support the multistage authentication framework. Primo et al. suggest creating two separate templates for smartphones, one for use while a phone is placed in a pocket and  the other for use while a phone is in a user's hand to improve the authentication accuracy of gait-based authentication systems. Next, the repeated verification component receives a stream of data and classifies that into either a genuine or an impostor categories, by using the template created during the enrollment process. The re-login component is typically invoked after a certain number of continuous false rejects, prompting the user to input some other credentials e.g. PIN, password etc. in order to verify their identity again. Finally, the template update component is responsible for updating template(s) after a certain period of time or after a drastic change in the environment\cite{ContinuousAuthJain}.\\
The proposed authentication system in this paper is a type of CA as it contains all of the above mentioned components. We assume that by using human activity recognition methods (see \cite{ActivityRecognition}\cite{EnergyEfficient}\cite{KwapiszActivity}), walking patterns can be accurately detected. The segment of data generated during the walking activity can be supplied to the proposed authentication system. The enrollment component in the proposed system consists of data collection, preprocessing, sliding window-based feature extraction, classifier training, and the computation of the user's specific equal error rate threshold. We used around two minutes of data for enrollment. Further investigation needs to be conducted about how the amount of training data affects the performance of the system and what the optimal size or duration of data for training should be. The repeated verification is done on a specified interval (2 to 4 seconds) based on the requirements and assuming the data is available continuously. The re-login authentication component is not integrated into our system; however, any conventional authentication system can be incorporated. We studied the impact of template updates by training on \textit{Session1} \textit{Phase2} data, updating the thresholds, and by testing on \textit{Session2} \textit{Phase2} data. We observed that, by updating the template we could maintain the actual performance of the system (i.e. accuracy of more then 95\%). For details, see Section \ref{sec:TemplateUpdate} and Table \ref{InterPhaseINterSession}. We could not make conclusions about what frequency (or duration) of update the template should be. We are planning to follow-up the data collection to investigate this problem further.
\subsection{Attack Scenarios}
\label{sec:ThreatModel}
A total of eight basic sources of attack are described in \cite{AttackSources} for a generic biometric system. The sources include, reproduction of biometric at the sensor (also known as mimicry attack), replay of stored (or stolen) biometric, overriding the feature extractor, snooping and tempering the feature vectors, overriding template matcher, tempering with or replacing stored templates (adopted in the movie Mission:impossible - Rogue Nation), channel attack, and decision override. The mimicry is an easy, unnoticeable, and practical method of attack on any biometric based authentication system, as it does not require any modification to the device or the authentication system. Therefore, biometric researchers focus more on mimicry attacks. Thus far there exists three kinds of mimicry attacks, in behavioral biometrics based authentication systems, namely, zero-effort, minimal-effort, and high-effort. In zero-effort mimicry attack, it is assumed that adversaries do not have access to the details of genuine user. Hence, imitators are arbitrarily chosen either from or outside of database. On the contrary, in minimal-effort mimicry attack, it is assumed that adversaries have access to the details of the genuine user. Hence, the imitators are chosen based on certain criteria such as similar physical characteristics. For example, to test a gait-based authentication mechanism, imitators who have similar height, weight and ethnicity should be used. The high-effort mimicry attack involves intensive training imitators to mimic behavior of the genuine user \cite{Gafurov}\cite{TreadmillAttack}\cite{SerwaddaRobotics}. For example, if an individual can be trained to type, swipe or walk like other individuals then the security provided by authentication systems based on these patterns will be of less use. In this paper, we evaluate the performance assuming the zero effort or random attack. However, we have recorded videos of individuals while collecting the data. We plan to evaluate our system by carrying out the minimal and high effort mimicry impostor attacks in the future.
\section{Data Collection} 
\label{sec:DataCollection}
\subsection{Procedure}
\begin{figure}[htp]
 \centering
 \includegraphics[width=2in, height=1.5in]{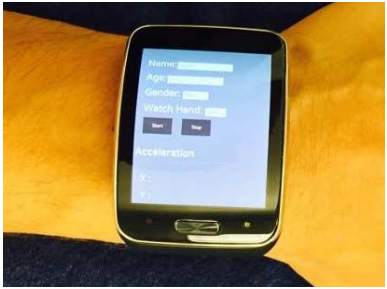}
 \caption{Samsung Gear S worn on the wrist of one of our participants.}
 \label{Watch}
\end{figure}

Following the approval from our university's institutional review board (IRB), we requested students, staff and faculty members of our university to participate in our data collection exercise. The participants were informed, prior to participation, that participation is completely voluntary and no compensation will be provided. A consent form, which briefly explained the nature of our study, was signed by each participant before we started the data collection process. Participants in our study were mostly graduate and undergraduate students, with the exception of a few faculty members and staff. In addition to collecting arm movement information through accelerometer and gyroscope sensors, we also collected \textit{age, gender, and WatchHand (left or right)} information of the participants (see Figure \ref{Watch}). The data collection was carried out in two phases (Phase1 and Phase2), each separated by at least three months. Each phase consists of two sessions (Session1 and Session2) separated by at least a ten minute time interval. The data collected during Session1 was used for \textit{training} the classifiers, and the Session2 data was used for \textit{testing} purposes in the \textbf{inter-session} testing setup. The data collected during Session1 of Phase1 was used to train the classifiers and the data collected during the Session1 and Session2 of Phase2 were used for testing under \textbf{inter-phase} testing setup. Forty subjects participated during Phase1. Twelve of them followed up in Phase2. Of the 40 subjects who participated in Phase1, 34 of them are between 20 and 30, four of them are between 30 and 35, and two of them are in their 50s. Ten of the subjects are female, while the rest are male. Of the twelve participants of Phase2, four of them are female and the rest are male, and all of them are between 20 and 30. No participants were given any specific instructions during the data collection except to wear the watch and walk as naturally as possible for around $\sim 2$ min.\\
\subsection{Application Development}
\label{subsec:ApplicationDevelopment}
We developed an application for the smartwatch (Samsung Galaxy Gear S) to record arm movements (acceleration and rotation) through the accelerometer and gyroscope sensors. We used the tizen-sdk version 2.3.63, which provides application programming interfaces (APIs) for accessing the sensor readings on the Ubuntu-64 bit platform. HTML and Javascript were used to create the graphical user interface. The user interface consists of four input fields to input \textit{userid, age, gender, and watchhand (left or right)}. The interface also had two buttons \textit{start} and \textit{stop} to control when data collection was to begin and end. The sampling rate for both the sensors, accelerometer and gyroscope was kept to 25Hz.
\subsection{Data Preprocessing}
\label{subsec:Procedure}
In order to remove the noise from the data before feature extraction, we used a simple (equally weighted) moving average technique, which is described as follows. Let $T(t)$ = ($x(t_1)$, $x(t_2)$, $x(t_3)$,...,$x(t_n)$) be the original data and $T'(t)$ = ($x'(t_1)$, $x'(t_2)$, $x'(t_3)$,...,$x'(t_{n-p})$) be the transformed data. Then $x'(t_i)$ are obtained as $1/p\times (x(t_{i})+x(t_{i-1})+...+x(t_{i-(p-1)}))$, where $p$ is the parameter to control the number of points taken at a time. We set $p$ to five in our experiment as it was able to remove the noise from the data without disturbing its characteristics.
\begin{figure*}[htp]
\centering
\begin{tabular}{cccc}
\subfigure[$A_x$ along the X-axis.]{\epsfig{file=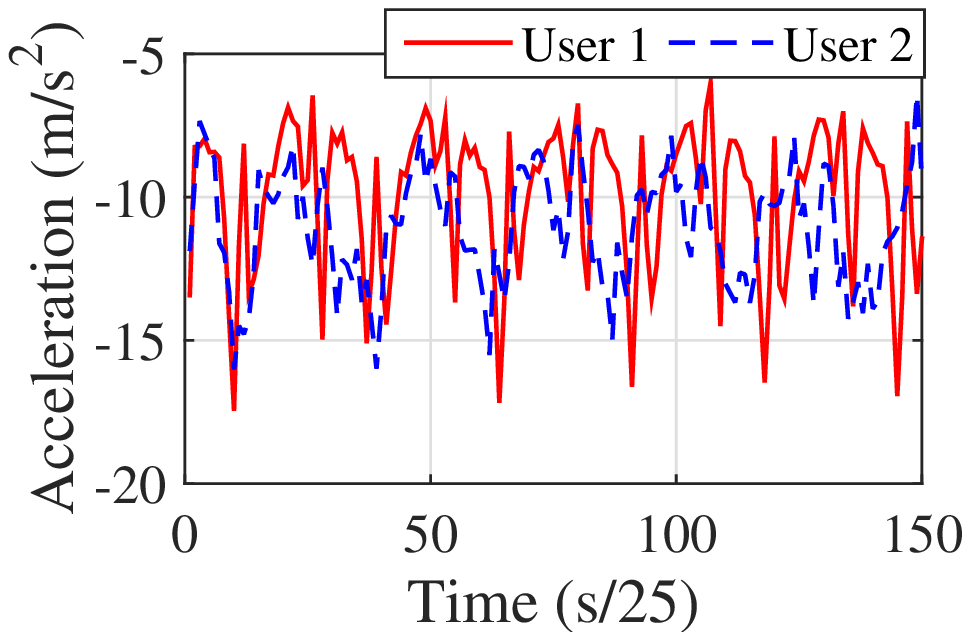,width=1.5in, height=0.8in}
\label{accx}}&
\subfigure[$A_x$ along the Y-axis.]{\epsfig{file=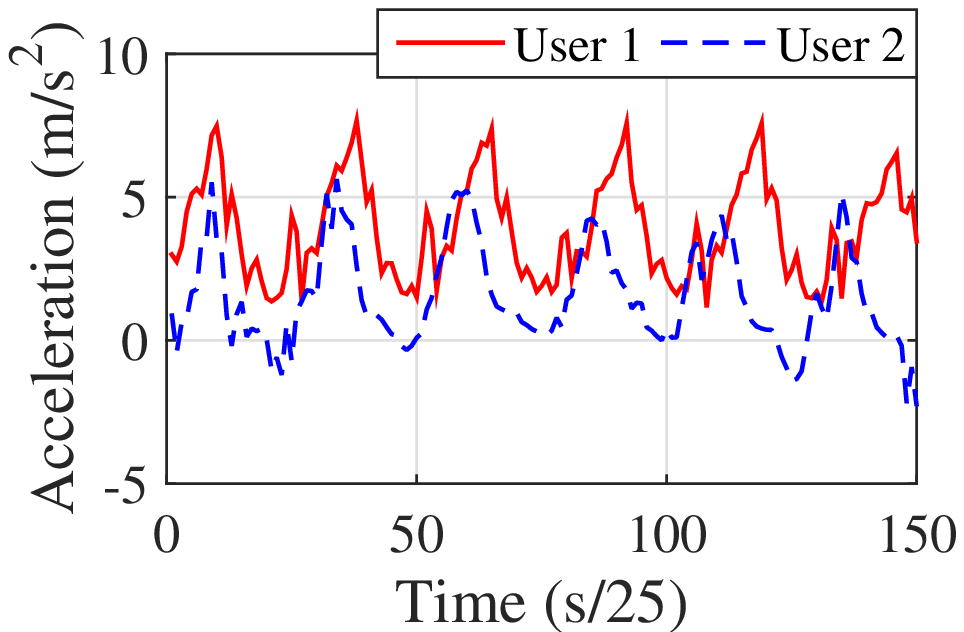,width=1.5in, height=0.8in}
\label{accy}}&
\subfigure[$A_x$ along the Z-axis.]{\epsfig{file=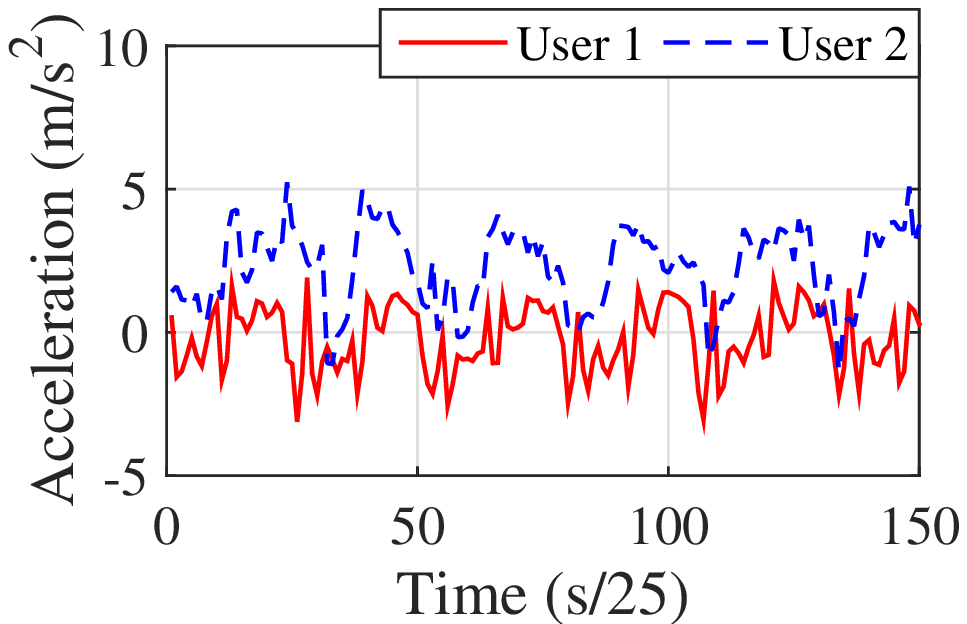,width=1.5in, height=0.8in}
\label{accz}}&
\subfigure[$A_m$ along the M-axis.]{\epsfig{file=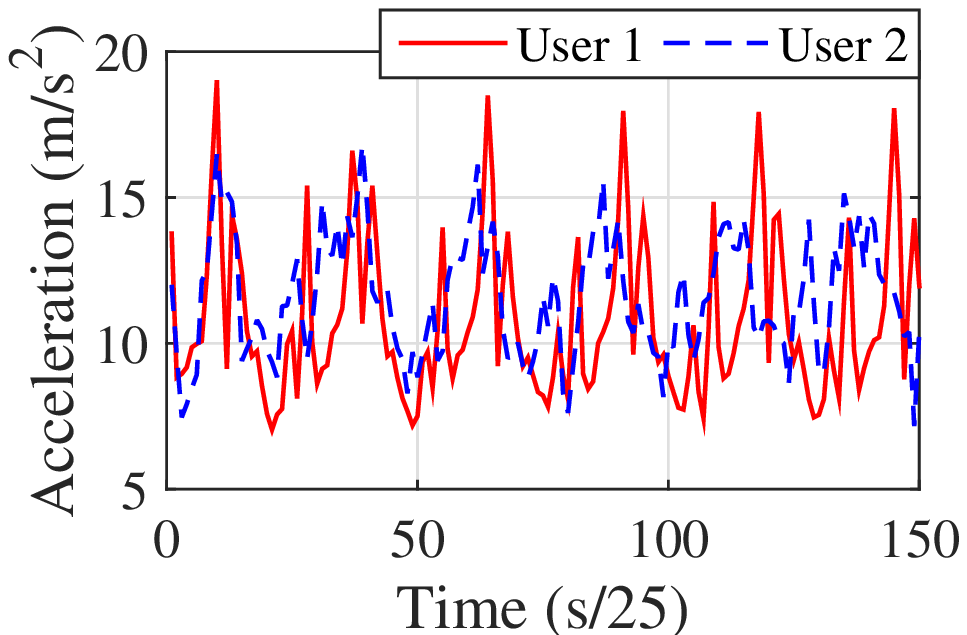,width=1.5in, height=0.8in}
\label{accm}}\\%
\subfigure[$R_x$ along the X-axis.]{\epsfig{file=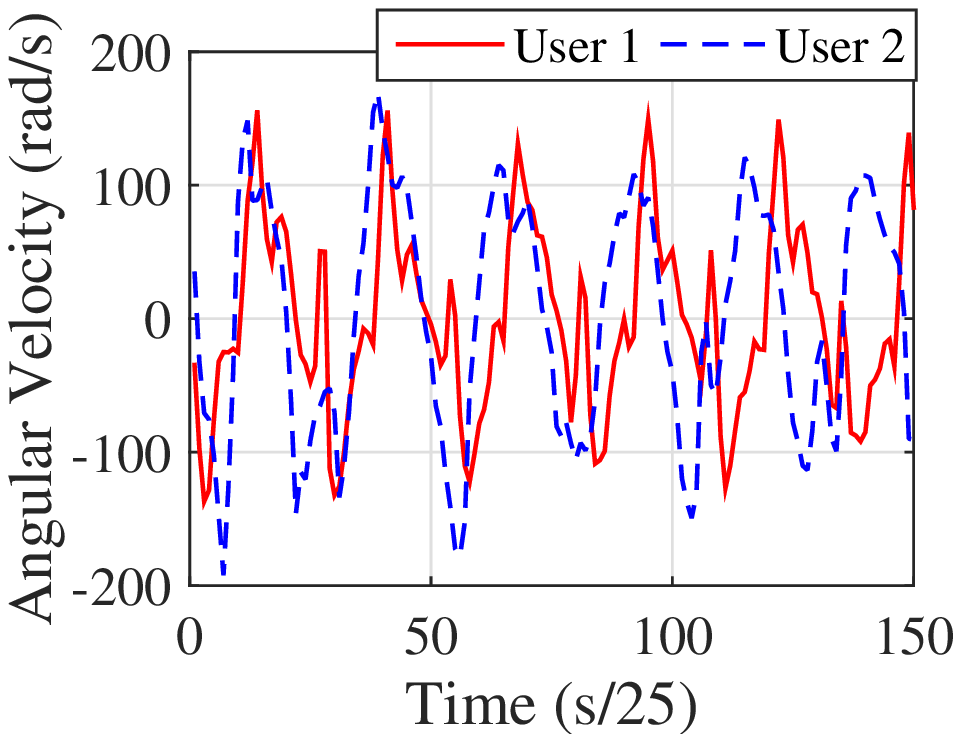,width=1.5in, height=0.8in}
\label{gyrox}}&
\subfigure[$R_y$ along the Y-axis.]{\epsfig{file=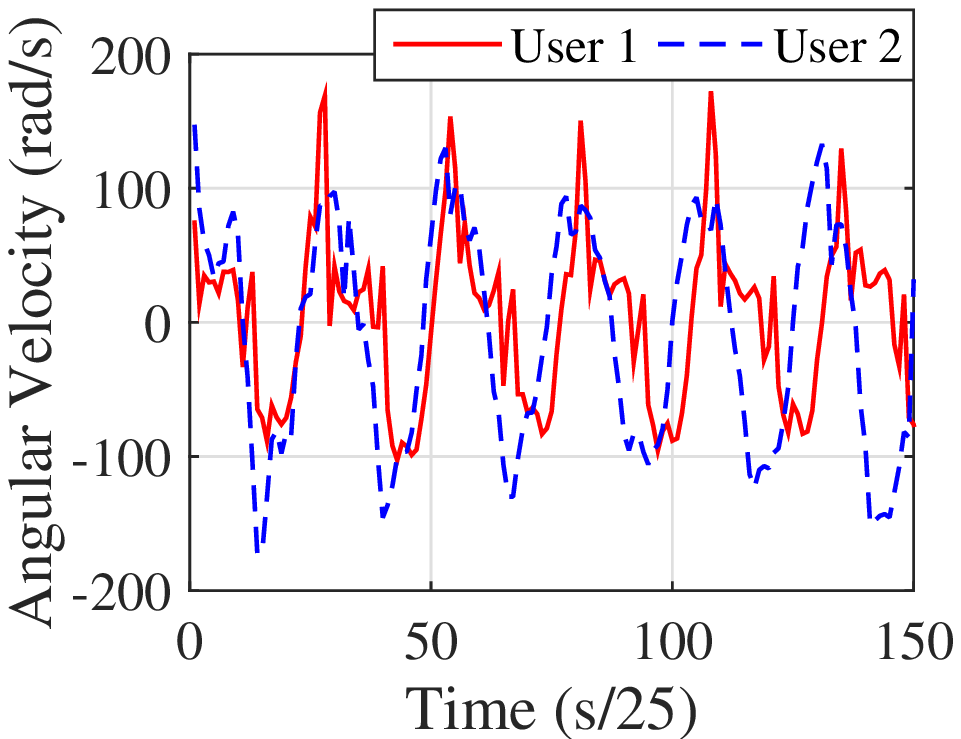,width=1.5in, height=0.8in}
\label{gyroy}}&
\subfigure[$R_z$ along the Z-axis.]{\epsfig{file=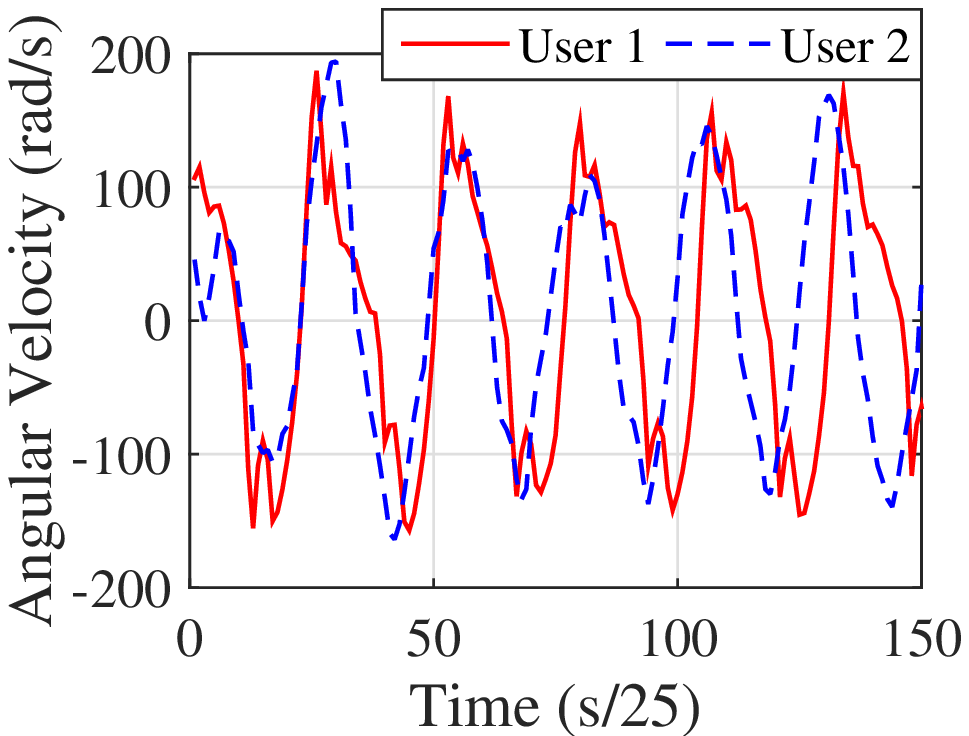,width=1.5in, height=0.8in}
\label{gyroz}}&
\subfigure[$R_m$ along the M-axis.]{\epsfig{file=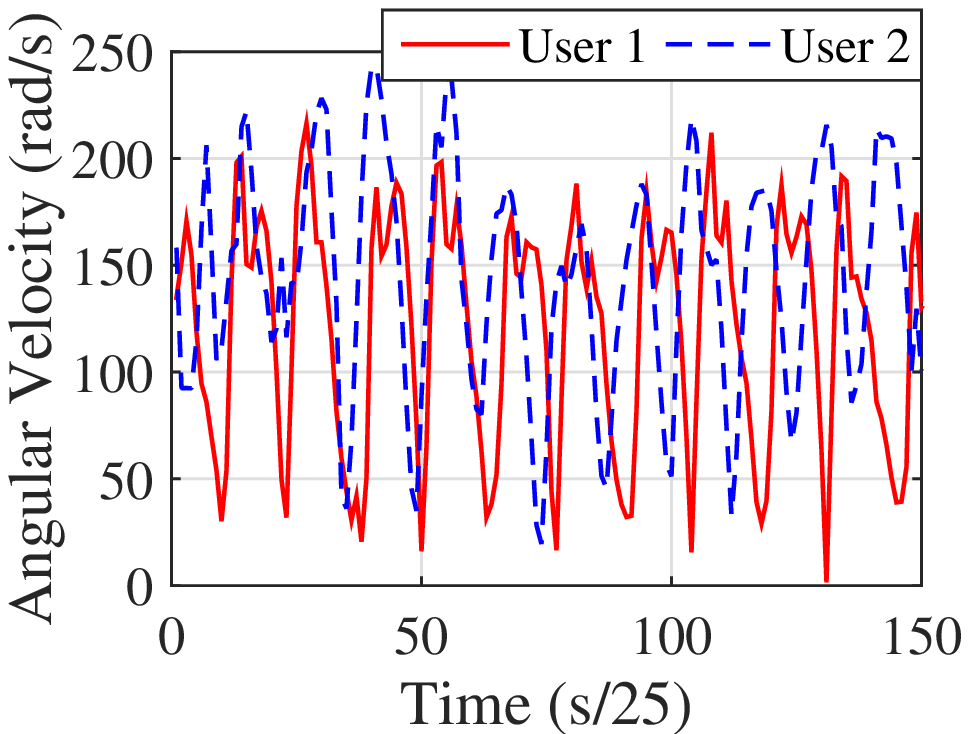,width=1.5in, height=0.8in}
\label{gyrom}}
\end{tabular}
\caption{Exploratory data analysis of segments of data taken from two arbitrarily selected users. A distinguishing trend between two arbitrary users can be observed.}
\label{TimeSeriesPlot}
\end{figure*}
\section{Feature Analysis}
\label{sec:FeatureAnalysis}
\subsection{Feature Extraction}
\begin{figure*}[htp]
  \centering
  \includegraphics[width=4in, height=0.9in]{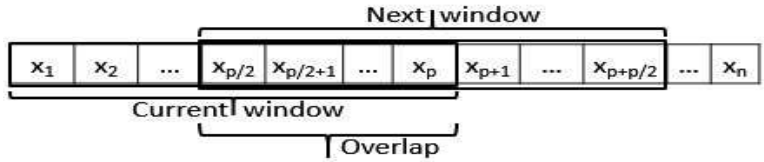}
  \caption{Sliding window based feature extraction}
  \label{SlidingWindow}
\end{figure*}

In order to observe whether the accelerometer and gyroscope readings vary significantly among the users, we carried out exploratory data analysis by plotting five to ten seconds of the data corresponding to each component of the acceleration $(a_x, a_y, a_z,$ and $a_m)$ and each component of the rotation $(r_x, r_y, r_z,$ and $r_m)$ for a few randomly selected users. Where $a_m$ and $r_m$ are defined as\\ $\sqrt{(a_x)^2 + (a_y)^2 + (a_z)^2}$ and $\sqrt{(r_x)^2 + (r_y)^2 + (r_z)^2}$  respectively. We observed clearly distinguishable cycles among each component of distinct users' data. We used a sliding window-based feature extraction mechanism in which the data stream is divided into several parts (windows) and each window overlaps with the previous one (see. Figure \ref{SlidingWindow}). The sliding (overlapping) windows-based segmentation of streaming data forms the basis for continuous authentication. In the authentication mode, a
segment of data is taken and features are extracted from it to obtain one feature vector. The feature vector is later fed to the trained classifier that outputs a score (see Section \ref{sec:ExperimentalDesign}). By comparing the obtained score with a pre-determined user specific threshold, the system decides whether it should accept or reject. The process loops for subsequent segments created from the forthcoming data.
\subsubsection{Length of the window and sliding interval}
\label{subsubsec:LengthOfWindow}
In order to decide what data stream length offers the most distinctiveness, we extracted a number of features from all eight components. The length of the window is referred to as the $W_{size}$ and the overlap as the $S_{interval}$. Our proposed authentication mechanism follows a continuous framework in which authentication decisions are given every few seconds. The $W_{size}$ and $S_{interval}$ are two critical parameters as they determine the time the system takes to give the first authentication decision and the time taken in the subsequent decisions respectively. Therefore, we performed an empirical analysis in order to find out the optimal setting for these parameters. We evaluated the performance of the acceleration and rotation based systems separately for ten different settings ($2,4,6,...,10$ seconds) of $W_{size}$ and for four different settings \textit{($W_{size}$, $W_{size}/2$, $4,$ and $2$ seconds)} of $S_{interval}$. We observed that the performance stabilized for $W_{size}$ beyond 250 (10 seconds). Since, we want to minimize the $W_{size}$, therefore we suggest to use any number of seconds between 8 to 12 as we observed lowest error rates in this range for majority of the classifiers. We used 10 seconds of windows in all of our experiments. Similarly, we observed that the performance at $S_{interval}=4$ seconds was relatively better compared to $S_{interval}=2$ seconds throughout, so we used $S_{interval}=4$ throughout the experiment.\\
With these settings for $W_{size}$ and $S_{interval}$, we extracted a total of 32 features ( = 8 unique features $\times$ 4 dimensions) from the data generated by arm acceleration and 44 features ( = 11 unique features $\times$ 4 dimensions) from the data generated from arm rotation (see Table \ref{ListOfTotalFeatures}).
\begin{table}[h]
\centering
\tbl{A list of features and their abbreviations used in this paper. We derived a total of 32 ( = $8$ (features) $\times$ $4$ (x, y, z, and m), from the accelerometer readings and 44 ( = $11$ (features) $\times$ $4$ (x, y, z, and m) from the gyroscope readings.}{
\begin{tabular}{|l|l|l|}
\hline
{\bf S.No.}       &{\bf Feature description}    & {\bf Abbrev.}\\ \hline
{\bf 1}           &{Average peak interval}      & API     \\ \hline
{\bf 2}           &{Bandpower}           		& BAP     \\ \hline
{\bf 3}           &{Energy}                     & ENG     \\ \hline
{\bf 4}           &{Median}               		& MED     \\ \hline
{\bf 5}           &{\# of peaks (only acc.)}         & NOP     \\ \hline
{\bf 6}           &{Range}	                    & RNG     \\ \hline
{\bf 7}           &{Median frequency}        	& MDF     \\ \hline
{\bf 8}           &{Spectral entropy}           & SPE       \\ \hline
{\bf 9}           &{Median stride time (only rot.)}  & MST     \\ \hline
{\bf 10}          &{\# of mid-swing points (only rot.)}	    & NMSP      \\ \hline
{\bf 11}          &{Mean rotation angle (only rot.)} & MRA     \\ \hline
{\bf 12}          &{Mean rotation rate (only rot.)}  & MRR    \\ \hline
\end{tabular}}
\label{ListOfTotalFeatures}
\end{table}
Let $W=(W_{x},W_{y},W_{z},$ and $W_{m})$ be the window, where $W_{x}$, $W_{y}$, $W_{z}$, and $W_{m}$ are four column vectors representing the rotation or acceleration along the x, y, z and m directions. As the $W_{size}$ is set to 10 seconds, each of the vectors contains 250 values as the sampling rate is 25 Hz. Let $W_{x}$ = ($x_i$, $x_{i+1}$, $x_{i+2}$,...,$x_j$). Then the required features can be computed as described below:
\begin{itemize}
\item \textbf{Average peak interval (API):} We identified the indices of peaks (local maxima) in $W_{x}$ by using the findpeaks function of MATLAB \cite{MATLAB2013a}.\\ Let $I = (i_j, i_{j+1}, i_{j+2},...,i_n)$ be a set of indices of peaks found in $W_{x}$. Then API = $\frac{1}{n-1}$ $\sum_{j=1}^{n-1} (i_{j+1}-i_j)$.
\item\textbf{Bandpower:} The bandpower of a signal is defined as the average power in the given frequency range. We computed this feature by using the $bandpower(W_{x}, S_{f}, F_{r})$ function of MATLAB. Where $W_{x}$ is the signal vector, $S_{f}$ is the sampling frequency that is $25$ Hz and $F_{r}$ is the frequency range, which is $[0, (S_{f})/2]$ in our experiments.
\item\textbf{Energy:} The energy of a signal vector $W_{x}$ is computed as \begin{displaymath} \int_{i}^{j} |Wx(t)^2|dt \end{displaymath}.
\item\textbf{Median:} To compute the median we sort the elements of $W_{x}$ and then compute $1/2 \times (x_{n/2}+x_{n/2+1})$ as the number of elements in $W_{x}$ is always 250 in our case and is even.
\item\textbf{Number of peaks (acceleration only):} We identified be the indices of peaks (local maxima) in $W_{x}$ by using the findpeaks function of MATLAB.\\ Let $I = (i_j, i_{j+1}, i_{j+2},...,i_n)$ be a set of indices of peaks found in $W_{x}$. Then the $length(I)$ provides the number of peaks.
\item\textbf{Range:} Range of a $W_{x}$ is computed as $max(W_{x})-min(W_{x})$.
\item\textbf{Median frequency:} A frequency that divides the power spectrum into two regions with equal amplitude is known as the median frequency. We compute the median frequency (MDF) by using the $medfreq(W_{x},S_f)$ of MATLAB R2015a, where $S_f$ is the sampling frequency.
\item\textbf{Spectral entropy:} Spectral entropy describes the complexity of the signal. It is directly proportional to the peak of the signal power spectrum and similar to Shannon's entropy. With the use of the power spectral density as a probability density. It is computed as\\ $\frac{1}{\log (N)} \sum P_i \log (P_i)$.
\item\textbf{The number of mid-swing points (rotation only):} A mid-swing (MS) point, is defined as the highest peak within a single stride. A stride or a cycle is defined as the curve formed by data points from one peak to another. To find the peaks, we used the \textit{findpeaks($W_{x}$, 'MinPeakDistance', p, 'MinPeakHeight', q)} function of MATLAB. We set p to 10 to avoid the nearby peaks which mostly occurred due to noise. For the q, we used 40, as we observed that peaks below forty were either noise or not useful (see Figure \ref{TimeSeriesPlot})\cite{ETelemed}.
\item\textbf{Median stride time (rotation only):} Initial contact (IC) points are defined as the first local minimum after the MS point in a single stride. Let $(c_1, c_2, c_3,...,c_q)$ be the indices of ICs in $W_{x}$ then the median stride time (MST) is defined as $median(c_2-c_1, c_3-c_2, c_4-c_3,...,c_q-c_{q-1})$.
\item\textbf{Mean rotation rate (rotation only):} Let $M=(\mu_1,\mu_2,\mu_3,...,\mu_r)$ be the vector containing the statistical mean of the absolute values between a consecutive pair of MS points in $W_{x}$. Then the mean rotation rate (MRR) is defined as $mean(M)$.
\item\textbf{Mean rotation angle (rotation only):} The mean rotation angle is basically a function of MST and MRR. It is computed as $(MRR \times MST)$.
\end{itemize}
\subsection{Feature Selection}
\label{sec:FeatureExtraction}
For better classification results, it is important to analyze the discriminability of these features as they directly affect the classification results. Therefore, we evaluated all 32 features (see Table \ref{ListOfTotalFeatures}) extracted from the accelerometer readings and 44 features (see Table \ref{ListOfTotalFeatures}) from the gyroscope readings by using two different methods, namely IGFR \cite{BenchmarkingAttributeSelectors} and CFSS \cite{CfsEvaluator}. We anticipated a few features to perform better in comparison to the others. For example, the rotation angle which is derived from two variables (i.e. mean rotation rate and stride time) that are independent of each other, believed to be a highly discriminative feature among different users. While the average number of cycles (or number of peaks) would be relatively less distinguishing as the number of steps taken by different individuals in 10 seconds of window does not vary much. This hypothesis is validated by using IGFR method that ranks features by how discriminative they are with respect to the class label. We saw that MRA\_Y and MRR\_Z were ranked among the top five best features under the ranked rotation feature list (see Table \ref{InfoGainTable}). However, classification algorithms that we have used in this paper use a set of features at a time. Therefore, we used CFSS which returns the best subset of features for distinguishing the users.
\subsubsection{Correlation Based Feature Subset Selection (CFSS)}
\label{Correlationbased}
\begin{figure*}[htp]
\centering
\begin{tabular}{c}
\subfigure[Acceleration based features.]{\epsfig{file=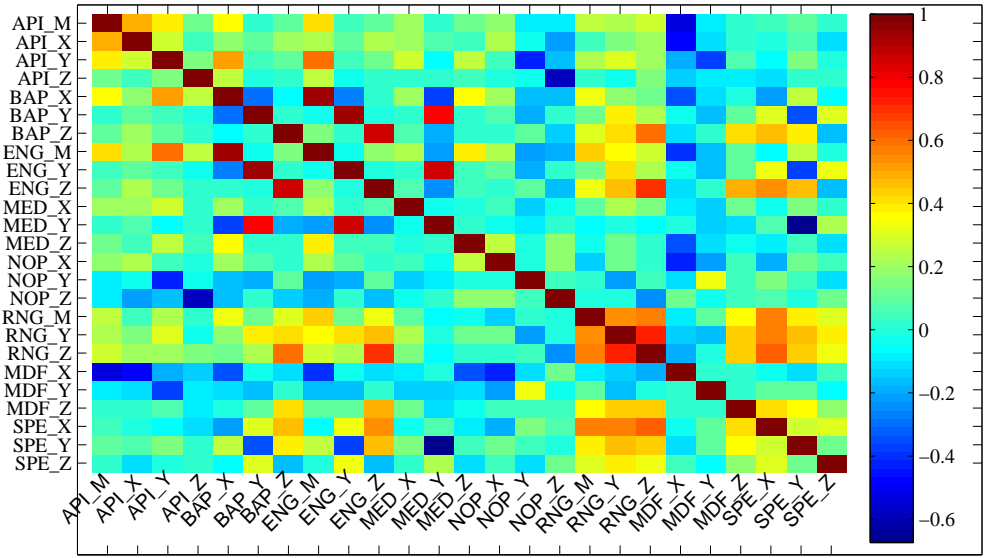,width=5.25in,height=3in}
\label{SelectedFeatureCorrelationAcc}}\\
\subfigure[Rotation based features.]{\epsfig{file=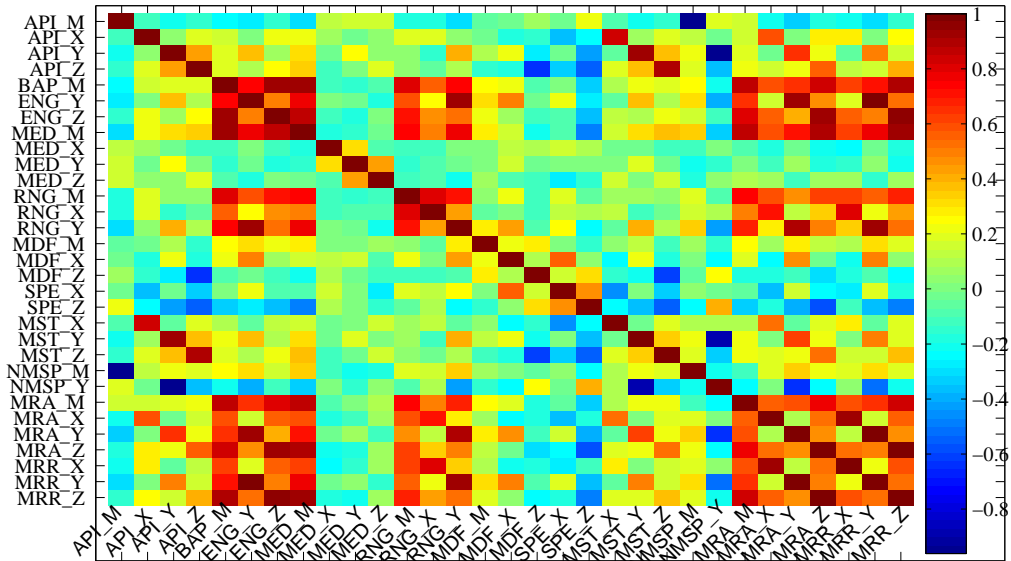,width=5.36in,height=3in}
\label{SelectedFeatureCorrelationGyro}}
\end{tabular}
\caption{Correlation among the subset of features selected from the set of acceleration and rotation based features by using CFSS method. As expected, we can see that most of the features are uncorrelated to each other.}
\label{SelectedFeatureCorrelationAccAndGyro}
\end{figure*}

We used CFSS implemented in Weka, with five possible search methods that use slightly different mechanisms to choose the resulting subset. These search methods included: the best first (BF), genetic search (GS), greedy stepwise (GRS), linear forward selection (LFS), and subset size forward selection (SSFS). For the 32 acceleration based features, the GS method chose 27 features, whereas the BF, GRS, LFS, and SSFS methods all selected the same 25 features. We computed the correlation among these features and present the results in Figure \ref{SelectedFeatureCorrelationAcc}. Similarly, of the 44 rotation-based features, the same 33 were selected by the BF and LFS, but GS, GRS, and SSFS selected only 31 features. Though the GRS and SSFS selected the same 31 features for their feature set, the GS method selected two different features. Because the different features that the GS method selected were ranked higher in the information gain table (see Table \ref{InfoGainTable}), we chose this as the best subset of the subsets of the 31 features. In order to see whether the CFSS worked, we computed the correlation among 25 accelerometer-based features (see Figure \ref{SelectedFeatureCorrelationAcc}) and also among 31 rotation-based features (see Figure \ref{SelectedFeatureCorrelationGyro}). We can observe that the correlations among the selected features are very low, which is seen as a good sign for classification purposes \cite{PhoneSwiping1}.\\
For feature-level fusion, we simply took a union of these separately selected accelerometer and rotation-based features (see Section \ref{subsec:FusionLevels}) which results in a set of 56 features. The set of fused features are referred to as separately the selected combined set (SSCS). However, one may wonder why we run feature selection separately, instead of combining the features and running feature selection on all of the features at once. Since, CFSS does not perform an exhaustive search over all the possible subsets, running the feature selection on a total of 76 features would be less likely to perform as exhaustively and accurately than feature selection run on the 32 features and the 44 features independently. The results of this would then be combined for another round of feature selection. To verify this concept, we applied the CFSS on the entire set of 76. The BF and GRS output the same 59 feature subset, the LFS and SSFS selected the same subset of 46 features, and the GS method selected 52 features. Next, when we evaluated the performances of the system built upon these feature subsets (59, 46, and 52), we found that the subset with 59 features performed slightly better when compared to the other two subsets (i.e. 46 and 52). We refer to the selected subset of features (59) as the selected set of total features (SSTF). Further, when we compared the performance of the SSCS and SSTF, the SSCS performed consistently better than the SSTF for all of the four classifiers used in our experiments. Therefore, we chose to use the SSCS in all of our feature-level fusion experiments. The heat map of the correlations among the SSCS features looks similar to Figures \ref{SelectedFeatureCorrelationAcc} and \ref{SelectedFeatureCorrelationGyro}.
\begin{table}[h]
\centering
\tbl{List of features ranked using the information gain attribute evaluator with the search method Ranker that ranks attributes by their individual evaluations. The attribute selection mode was the $10$-fold cross validation. Hence, we obtained the average information gain and the standard deviation. }{
\begin{tabular}{|l|l|l|l|l|}
\hline
\multirow{2}{*}{{\bf Rank}} & \multicolumn{2}{l|}{{\bf Acc. Features}} & \multicolumn{2}{l|}{{\bf Rot. Features}} \\ \cline{2-5}
                            & {\bf Feature}         & {\bf Info Gain}         & {\bf Feature}        & {\bf Info Gain}        \\ \hline
{\bf 1}                     & MED\_Y                & 2.50  (0.05)          & ENG\_Z            & 2.21 (0.03)          \\ \hline
{\bf 2}                     & ENG\_X             & 2.26  (0.05)          & MED\_M               & 2.16 (0.07)          \\ \hline
{\bf 3}                     & ENG\_M             & 2.24  (0.05)          & MRR\_M               & 2.17 (0.09)          \\ \hline
{\bf 4}                     & MED\_Z                & 2.21  (0.05)           & ENG\_M            & 2.15 (0.06)          \\ \hline
{\bf 5}                     & MED\_X                & 2.13  (0.03)          & MRR\_Z               & 2.14 (0.08)          \\ \hline
{\bf 6}                     & ENG\_Y             & 2.13  (0.08)           & ENG\_Y            & 2.12 (0.05)          \\ \hline
{\bf 7}                     & BAP\_X                 & 2.11  (0.03)           & RNG\_Y             & 1.99 (0.06)          \\ \hline
{\bf 8}                     & BAP\_M                 & 2.00  (0.08)          & MRA\_Z               & 1.99 (0.04)          \\ \hline
{\bf 9}                     & BAP\_Y                 & 1.96  (0.06)          & BAP\_M                & 1.96 (0.06)          \\ \hline
{\bf 10}                    & ENG\_Y             & 1.84  (0.14)          & MRR\_Y               & 1.94 (0.05)           \\ \hline
{\bf 11}                    & MED\_M                & 1.87  (0.02)           & RNG\_Z             & 1.93 (0.05)          \\ \hline
{\bf 12}                    & SPE\_Y                & 1.64  (0.06)          & BAP\_Z                & 1.90 (0.07)           \\ \hline
{\bf 13}                    & SPE\_X                & 1.60  (0.08)          & BAP\_Y                & 1.84 (0.03)          \\ \hline
{\bf 14}                    & BAP\_Z                 & 1.52  (0.03)           & MRA\_Y               & 1.78 (0.04)          \\ \hline
{\bf 15}                    & SPE\_M                & 1.42  (0.02)          & RNG\_M             & 1.73 (0.07)          \\ \hline
{\bf 16}                    & RNG\_Z              & 1.36  (0.11)          & MDF\_M               & 1.61 (0.06)          \\ \hline
{\bf 17}                    & RNG\_M              & 1.32  (0.02)          & MRR\_X               & 1.61 (0.13)          \\ \hline
{\bf 18}                    & MDF\_Y                & 1.32  (0.07)          & MRA\_M               & 1.57 (0.1)            \\ \hline
{\bf 19}                    & RNG\_X              & 1.27  (0.25)          & MRA\_X               & 1.44 (0.08)          \\ \hline
{\bf 20}                    & NOP\_Z                & 1.28  (0.07)           & MDF\_X               & 1.42 (0.05)           \\ \hline
{\bf 21}                    & MDF\_X                & 1.22  (0.01)          & MST\_Z               & 1.38 (0.05)          \\ \hline
{\bf 22}                    & MDF\_M                & 1.18  (0.10)          & ENG\_X            & 1.38 (0.09)           \\ \hline
{\bf 23}                    & NOP\_X                & 1.18  (0.13)          & MDF\_Z               & 1.29 (0.02)          \\ \hline
{\bf 24}                    & API\_M                & 1.10  (0.03)          & MST\_X               & 1.24 (0.06)           \\ \hline
{\bf 25}                    & API\_X                & 1.05  (0.13)          & API\_Z               & 1.17 (0.29)          \\ \hline
{\bf 26}                    & RNG\_Y              & 1.02  (0.23)          & API\_Y               & 1.22 (0.04)          \\ \hline
{\bf 27}                    & API\_Y                & 1.01  (0.11)          & BAP\_X                & 1.23 (0.09)            \\ \hline
{\bf 28}                    & NOP\_M                & 0.98  (0.16)          & MST\_Y               & 1.19 (0.08)          \\ \hline
{\bf 29}                    & MDF\_Z                & 0.90  (0.08)          & MDF\_Y               & 1.05 (0.04)          \\ \hline
{\bf 30}                    & API\_Z                & 0.83  (0.03)          & API\_M               & 1.01 (0.16)            \\ \hline
{\bf 31}                    & NOP\_Y                & 0.81  (0.01)           & RNG\_X             & 0.96 (0.04)          \\ \hline
{\bf 32}                    & SPE\_Z                & 0.73  (0.04)          & SPE\_Z               & 0.86 (0.04)          \\ \hline
\end{tabular}}
\label{InfoGainTable}
\end{table}
\subsubsection{Information Gain Feature Ranking Method (IGFR)}
\label{InfoGain}
As mentioned above, the CFSS method selects the best subset following the philosophy that, the selected features should be highly correlated with the class attribute in addition to being the least correlated with other features in the subset. The CFSS method, however, fails to provide any comparison between the features. Therefore, we used IGFR \cite{BenchmarkingAttributeSelectors}, which provides a measure of discriminability of each of the features separately and ranks them accordingly. We evaluated all 32 acceleration-based features by using the IGFR method. The features and their information gain are presented in Table \ref{InfoGainTable}. Interestingly, when observed we found that, of the 25 acceleration-based features that were selected by the CFSS, 20 were ranked in the top 25 features in the information gain table (see Figure \ref{SelectedFeatureCorrelationAcc} and Table \ref{InfoGainTable}). Similarly, we used IGFR methods on the 44 rotation-based features and ranked them according to their information gain; we found that, of the 31 rotation-based features selected by the CFSS, 22 were ranked within the top 30 features (see Figure \ref{SelectedFeatureCorrelationGyro} and Table \ref{InfoGainTable}). It will be interesting to see the tradeoff between the number of features used for classification and the performance. We plan to carry out this analysis in our future work.
\section{Performance Evaluation}
\label{sec:ExperimentalDesign}
\subsection{Classification Algorithms}Different researchers have used different classification algorithms for building authentication systems based on acceleration captured by smartphones. For example, Yang et al. \cite{HMMSmartWatch} used histogram and dynamic time warping methods, Primo et al. \cite{ABENA} used Logistic Regression, Zhong et al. \cite{GDIBasedGBAS} used gait dynamics images, Nickel et al. \cite{kNNForGait} used k-NN, and Kumar et al. \cite{TreadmillAttack} used Bayes Network, Logistic Regression, Neural Network, Random Forest, and SVMs. Along the same line, we chose to use k-NN with Euclidean distance and ten nearest neighbors \cite{kNNEuc}, and Random Forest with one thousand trees \cite{RandomForest}. These two algorithms are commonly known as weighted neighborhood schemes, and are comparatively faster in decision-making. We also used Logistic Regression \cite{LogisticRegression} and Neural Network (i.e. Multilayer Perceptrons) \cite{MultilayerPerceptrons}, as both are frequently used classifiers in this area \cite{FordhamSmartWatch}. The Multilayer Perceptrons was observably slow in training and testing, and also did not perform better than the other classifiers.
\begin{figure*}[htp]
\centering
\begin{tabular}{cc}
\subfigure[Distribution of user-specific EER thresholds for acceleration based system.]{\epsfig{file=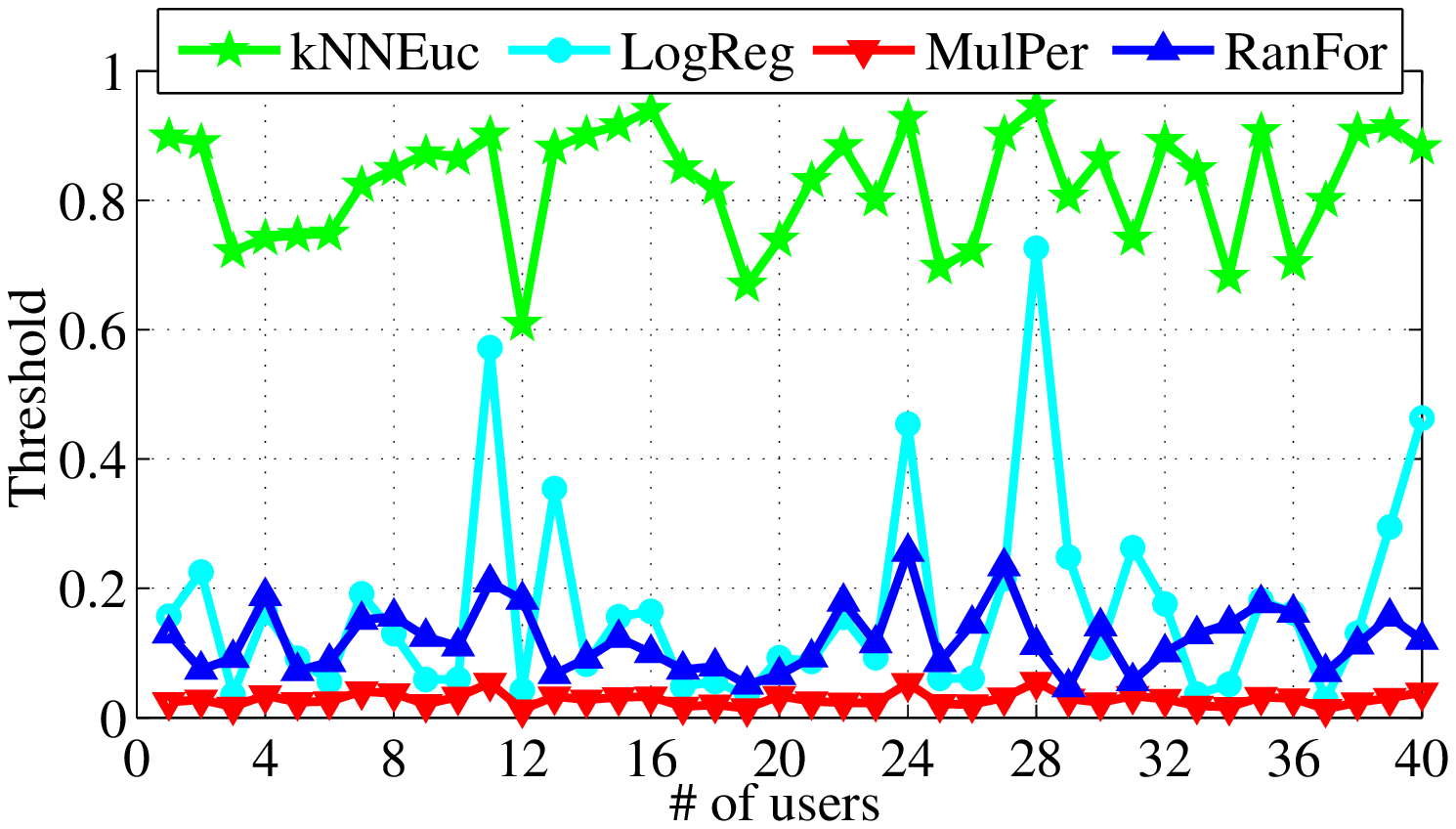,width=3.25in, height=1.2in}
\label{UserThresholdDistributionAcc1}}
\subfigure[Distribution of user-specific EER thresholds for rotation based system.]{\epsfig{file=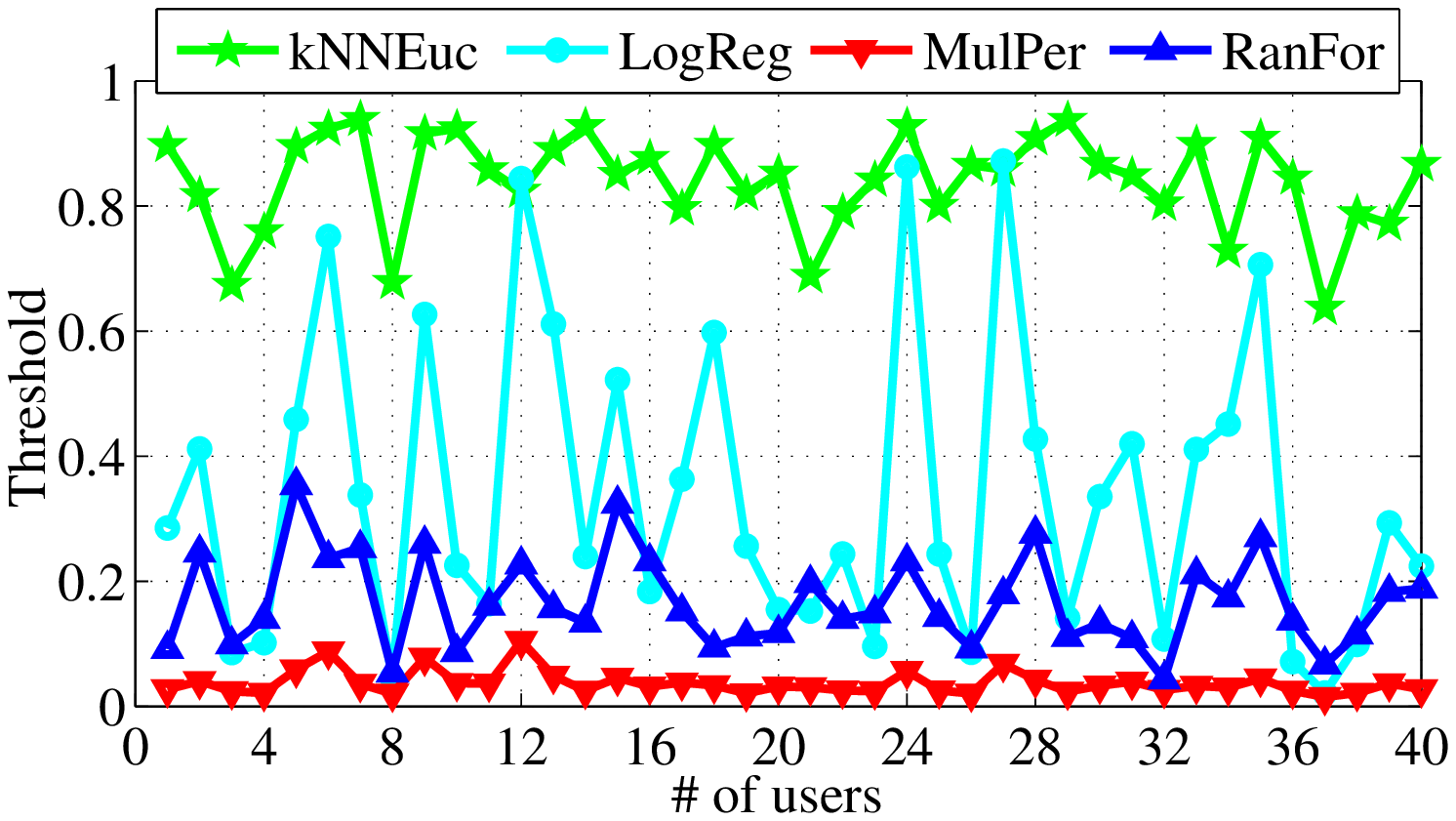,width=3.25in, height=1.2in}
\label{UserThresholdDistributionGyro1}}
\end{tabular}
\caption{The distribution of user-specific dynamic equal error rate thresholds. These figures suggest that a global threshold could have downgraded the performance of the authentication systems.}
\label{UserLevelThreshold}
\end{figure*}
\subsection{Training and Testing Methods}
\subsubsection{Training the classifiers}
We used a total of four state-of-the-art classifiers, all of them requiring knowledge of both genuine and impostor classes in order to be trained. In other words, the training set consists of both genuine and impostor samples. The genuine and impostor samples consist of a set of feature vectors that are created by extracting features from overlapping windows. The overlapping windows are obtained by segmenting the raw data (see Figure \ref{SlidingWindow}). These samples are created for each user separately. We refer to the user as the \textbf{candidate} user for which the samples are being created, whereas the remaining users are referred to as \textbf{other} user. The process of creating feature vectors for genuine and imposter samples is the same. The only difference is, the raw data used for creating them. For instance, the feature vectors of genuine samples are created by using the \textbf{candidate} user's data, whereas the feature vectors of impostor samples are created by using the other user's data. To create impostor samples, we investigated three options. First, we took genuine samples of every other user. Second, we took a fixed number of genuine samples from every other user. Third, we selected a fixed number of genuine samples from a fixed number of randomly selected users from the set of \textbf{other} users.\\
We decided to use the second option, i.e., the fixed number (four consecutive) of samples from every \textbf{other} user. The main reason behind choosing this option is that it gives relatively more samples for training and testing compared to the third option.  Additionally, it avoids the class imbalance problem to some extent which would have been worse compared to the first option. However, we observed that, even with the second option we had more feature vectors (39 ( \# of other users) * 4 ( \# of feature vector taken from each) = 156 feature vectors) in the impostor samples compared to the genuine samples (15-20) for a user. One feature vector is created by using ten seconds of window and we had $\sim$ 2 min of data for each user. By keeping the overlap of four seconds between the consecutive windows, we had around 15-20 feature vectors created from $\sim$ 2 min of data for each user. The difference in the number of feature vectors of the genuine (15-20) and impostor (156) samples is still big and can potentially pose a class imbalance problem. To overcome this issue, we again had two alternatives, either over-sampling the genuine users or under-sampling the impostor samples. We opted to over sample the genuine samples by using bootstrapping (repeating the uniform-randomly picked instances) to make the number of genuine and impostor samples equal. As a result, the training set contains an equal number (156) of features vectors for both impostor and genuine classes for each user. The training set is fed to the classifiers to create an independent model (template or profile) for each user.\\
For inter-session setup, the training sets were created from data collected during Phase1 Session1 ($P_1S_1$) for Phase1 inter-session whereas data collected during Phase2 Session1 ($P_2S_1$) was used for Phase2 inter-session (see Table \ref{InterPhaseINterSession}). Similarly, for inter-phase setup, the training set was created from data collected during Phase1 Session1 ($P_1S_1$).
\subsubsection{Genuine and Impostor Testing}
\begin{figure*}[htp]
\centering
\begin{tabular}{cc}
\subfigure[Average DFAR.]{\epsfig{file=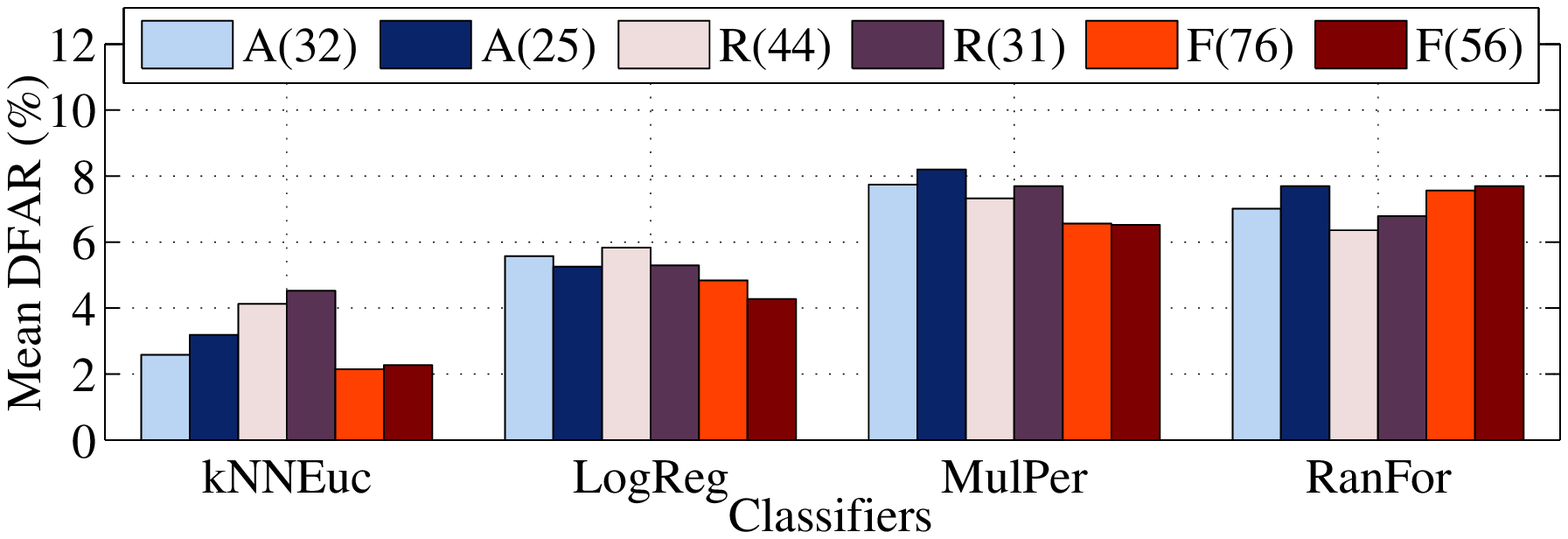,width=3.2in, height=1.2in}
\label{PerformanceCompFAR}}&
\subfigure[Average DFRR.]{\epsfig{file=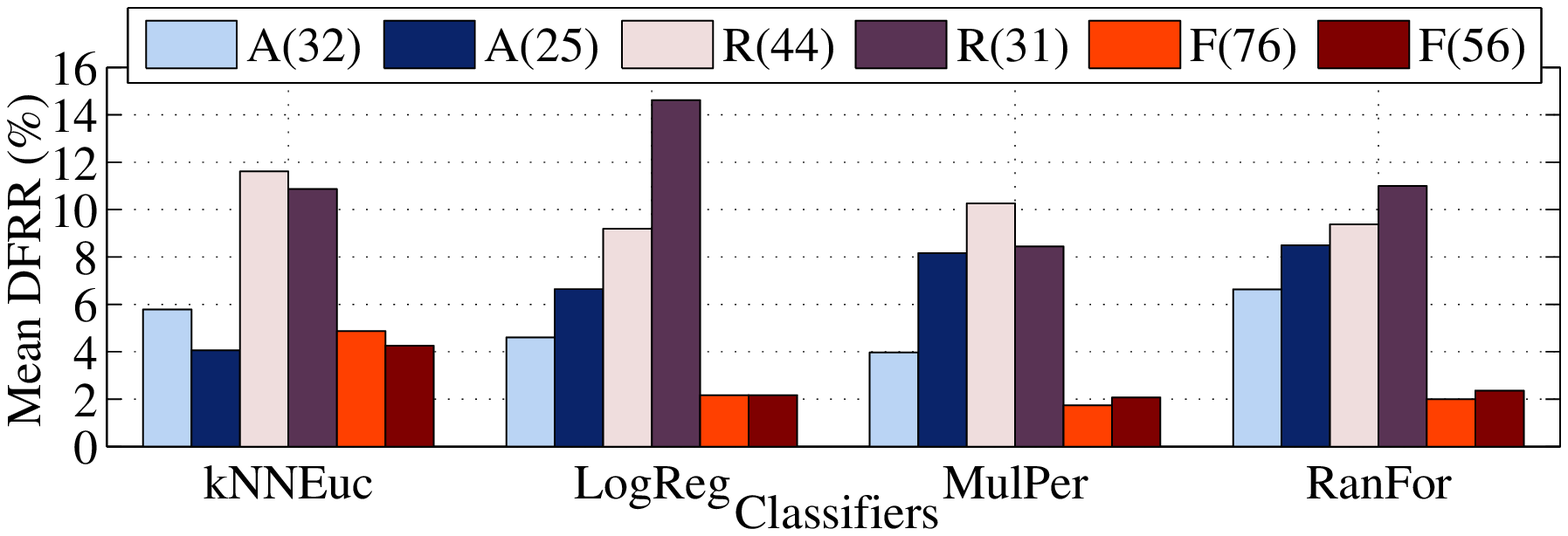,width=3.2in, height=1.2in}
\label{PerformanceCompFRR}}\\
\subfigure[Average Dynamic Accuracy.]{\epsfig{file=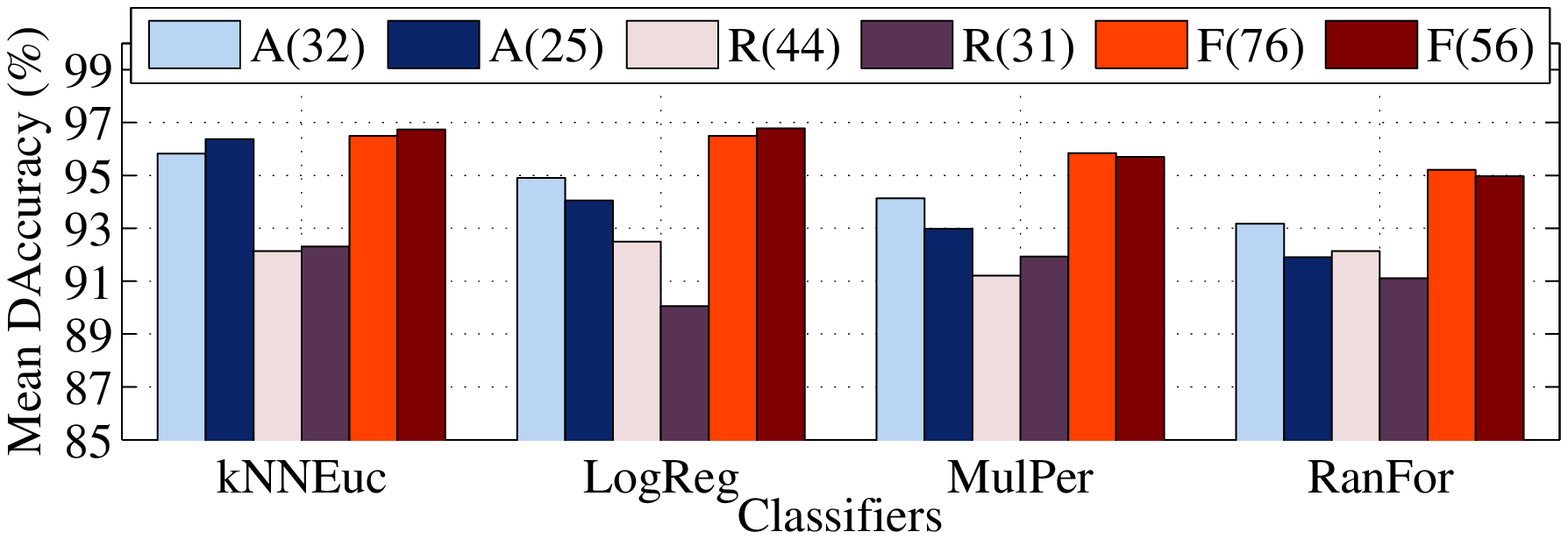,width=3.2in, height=1.2in}
\label{PerformanceCompAccuracy}}&
\subfigure[Scalability of the system built only upon acceleration data generated from hand movements.]{\epsfig{file=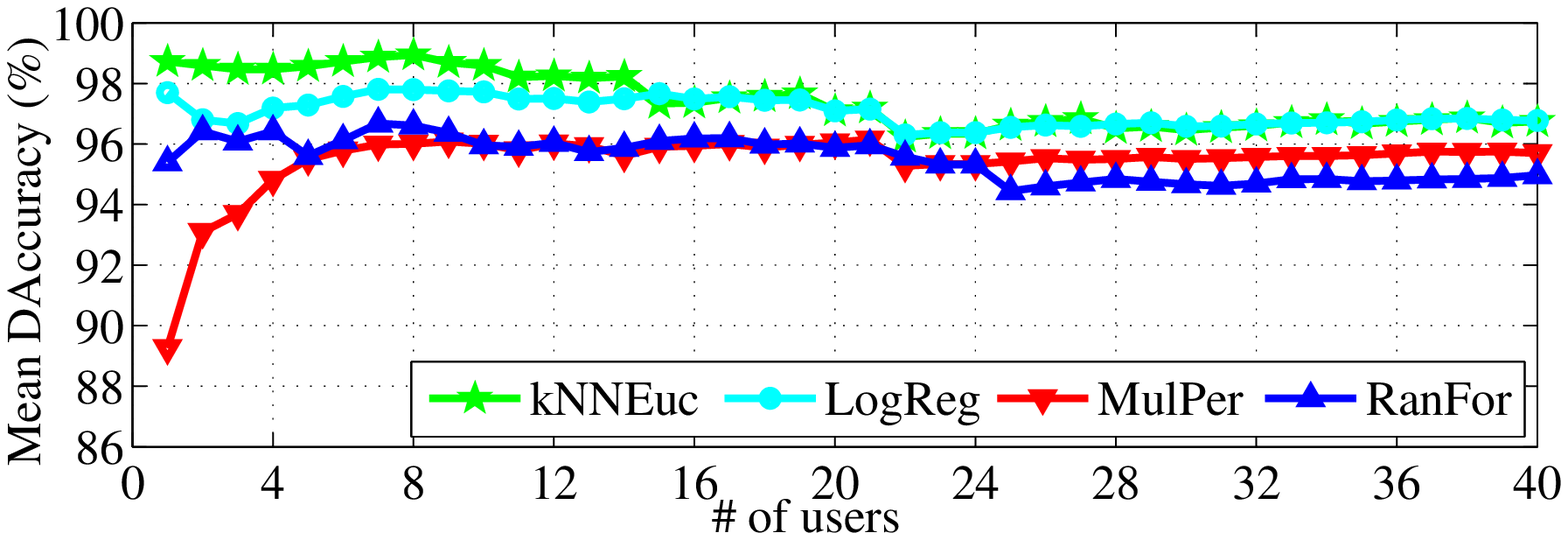,width=3.2in, height=1.2in}
\label{ScalabilityAcc}}\\
\subfigure[Scalability of the system built only upon rotation data generated from hand movements.]{\epsfig{file=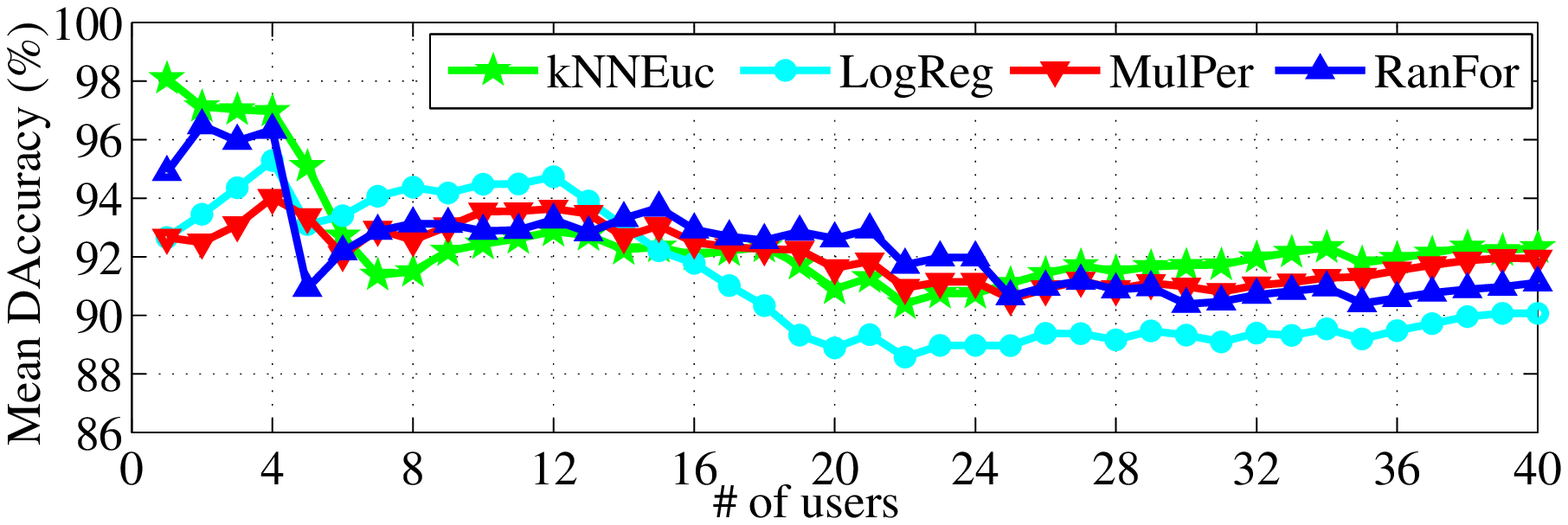,width=3.2in, height=1.2in}
\label{ScalabilityGyro}}&
\subfigure[Scalability of the system built upon the fusion of acceleration and rotation generated from hand movements.]{\epsfig{file=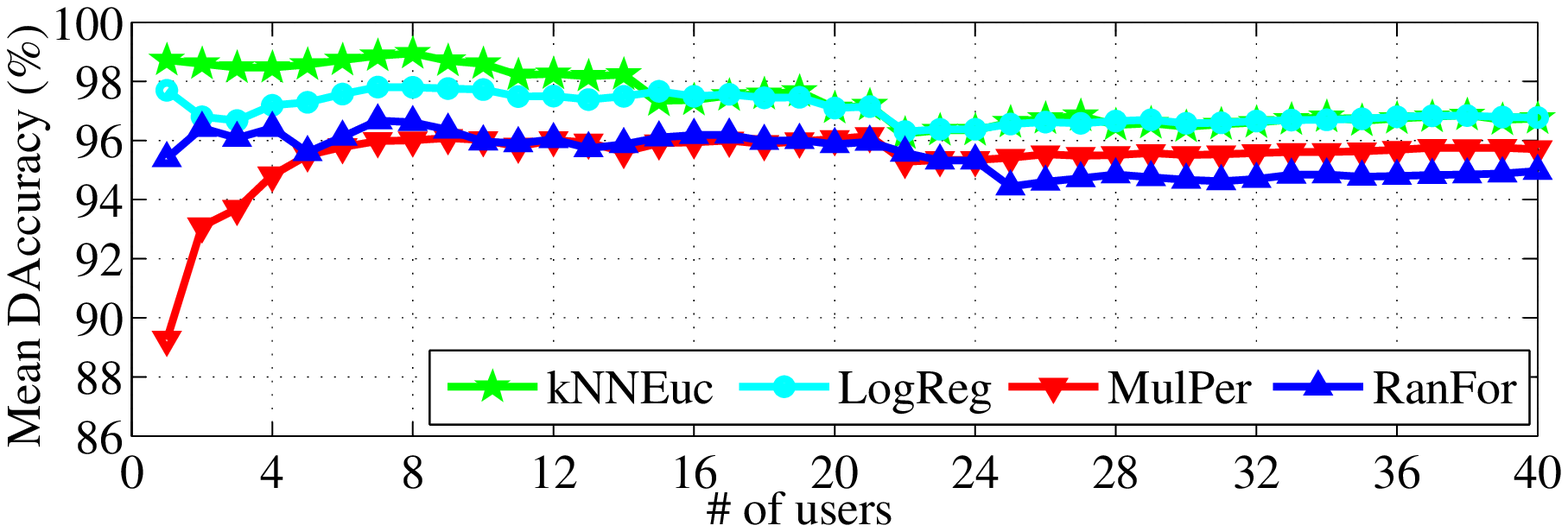,width=3.2in, height=1.2in}
\label{ScalabilityFusion}}
\end{tabular}
\caption{Figures \ref{PerformanceCompFAR}, \ref{PerformanceCompFRR}, and \ref{PerformanceCompAccuracy}, show the impact of feature selection and FLF on the average performance (DFAR, DFRR, and dynamic accuracy) of the authentication system. In this figure, acceleration, rotation and fusion are abbreviated as A, R and F respectively. The numbers of features used for corresponding sensors are given inside parentheses. For example, A(32) means the performance is evaluated by using a total of 32 acceleration-based features. Moreover, Figures \ref{ScalabilityAcc}, \ref{ScalabilityGyro}, and \ref{ScalabilityFusion} show the scalability of the systems built upon the acceleration, gyroscope, and fusion of these two respectively.}
\label{PerformanceComparision}
\end{figure*}

Once the models are created for each user by using the training set, we test the model by using the genuine and impostor samples separately, created from data designated for testing. The testing data depends upon the testing setup. For example, for inter-session testing setup, the testing data comes from either Phase1 Session2 or Phase2 Session2. Whereas, for inter-phase testing setup, the testing data comes either from Phase2 Session1 or Phase2 Session2. The classifiers output probabilities of belonging to the genuine class for every single vector supplied to them. The probabilities obtained for the feature vectors belonging to the genuine class are referred to as the genuine score, whereas the probabilities obtained for feature vectors belonging to the impostor class are referred to as the impostor scores in the rest of this paper. For each user, we obtained separate vectors of genuine and impostor scores. Let $G_{s}=\{g_{1}, g_{2}, g_{3},..., g_{m}\}$ and $I_{s}=\{i_{1}, i_{2}, i_{3},...,i_{n}\}$ be the vectors of the genuine and impostor scores, respectively. These scores were obtained for the feature vectors created from the successive overlapping windows, but are considered as independent scores that make an authentication decision. Successive authentication decisions are made from these scores by using user-specific equal error rate (EER) thresholds. The user-specific EER thresholds are computed during the training period. To create the user-specific threshold, we used the genuine and impostor scores obtained by using the training data as test data. Figure \ref{UserLevelThreshold} shows equal error thresholds obtained for each user. We observed that the threshold for each user varies significantly, we opted to use user specific thresholds instead of a global threshold. Moreover, other researchers have also suggested that user-specific thresholds not only improve the performance, but also increase the robustness of the system against spoof attacks \cite{UserSpecificSpoofProof}\cite{UserSpecificJain}.\\
Since in our system, acceptance or rejection decisions are continuously made over a certain period of time we use the continuous authentication metrics DFAR, DFRR and dynamic accuracy \cite{BiometricsMetrics}\cite{BiometricsMetricsReport} to evaluate the overall performance of our system. For each user, we computed the DFAR, DFRR and dynamic accuracy by using the genuine (and imposter scores) and the specific equal error rate (EER) threshold computed for each user during the enrollment process. We report the average DFAR, DFRR, and dynamic accuracy, computed over a total of 40 users (e.g. see Figure \ref{PerformanceComparision}).
\subsection{Fusion of sensor readings} The fusion of acceleration and rotation is feasible at three different levels namely, at the data level, feature level, and score level. In this paper, we only investigate the feature and score level fusion. We plan to explore the data level fusion in future. These different levels of fusion proved to have varying levels of success in improving overall performance of the system and adding an extra layer of security against imitation attacks on authentication systems which are built by using acceleration only (see \cite{TreadmillAttack}).
\begin{figure*}[htp]
\centering
\begin{tabular}{cc}
\subfigure[FAR]{\epsfig{file=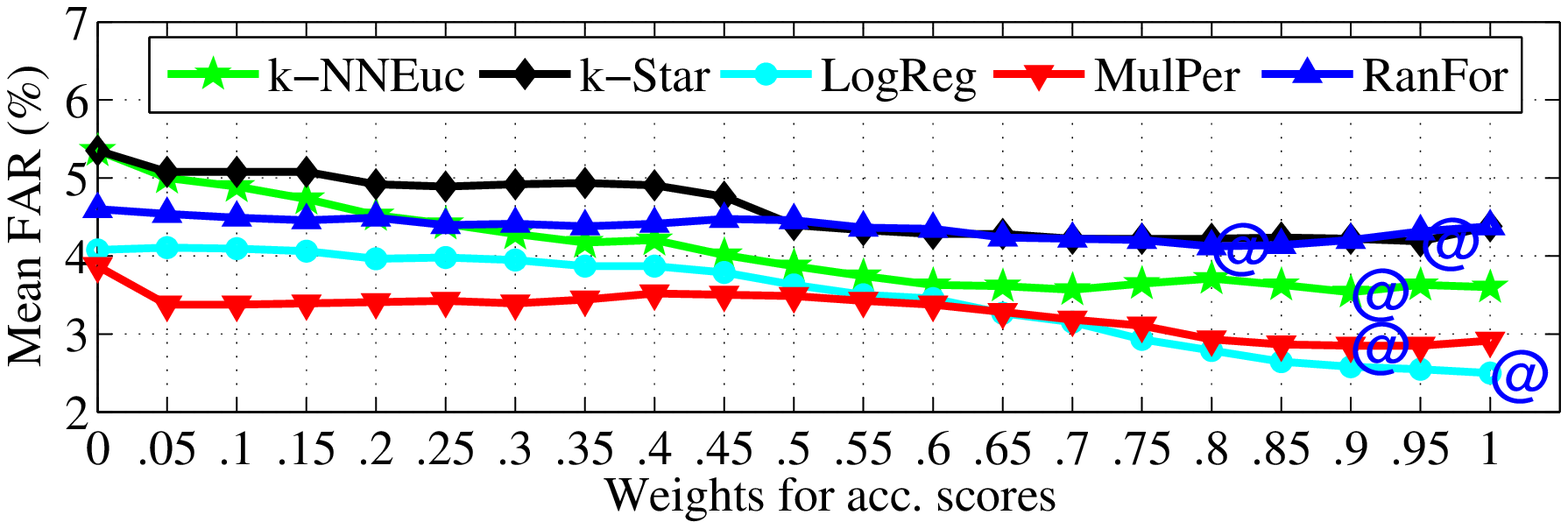,width=3.25in,height=1.4in}
\label{SLFFAR8}}
\subfigure[FRR]{\epsfig{file=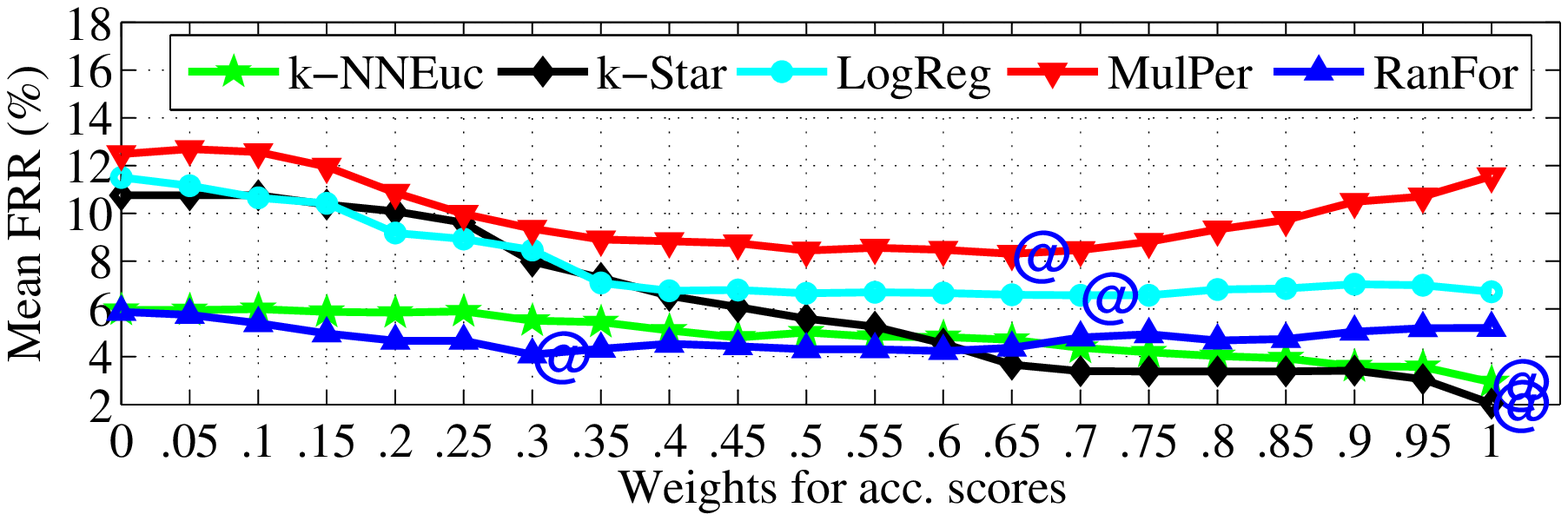,width=3.25in,height=1.4in}
\label{SLFFRR8}}
\end{tabular}
\caption{Impact of SLF on average FAR and FRR computed over all 40 users of the dataset. The weights along the x-axis are the
weights assigned to the acceleration based scores whereas the weights for rotation-based are computed as $(1-W_a)$.The at the rate symbols (@) show the best combination of weights for fusion.}
\label{SelectedFeatureCorrelationAccAdGyro}
\end{figure*}
\subsubsection{Feature level fusion}\label{subsec:FusionLevels}
Fusion at the feature level is proven to be effective in multi-modal biometrics systems. Therefore, we carried out a feature level fusion (FLF) considering acceleration (32) as one and rotation (44) as the other modality. As mentioned in Subsection \ref{sec:FeatureExtraction}, we evaluated these features separately and selected 25 and 31 features respectively. Further, to fuse them we applied the concatenation method. For example, let $A = \left\{f_{1},f_{2},...,f_{25}\right\}$ be the subset of features selected from the acceleration data and $G = \left\{g_{1},g_{2},...,g_{31}\right\}$ be the subsets of features selected from the rotation data, then the fused feature set can be given as $S = A \cup G = \left\{f_{1},f_{2},..,f_{25},g_{1},g_{2},...,g_{31}\right\}$ \cite{FLF1} \cite{FLF2}. In order to demonstrate the impact of the FLF on the performance, we evaluated three variations of the authentication system: first, by using only an acceleration-based system, second, by using only a rotation-based system, and third, by using the FLF-based system. In addition, to show the effect of feature selection, we evaluated these three systems with and without feature selection. The performance of all six systems is presented in Figure \ref{PerformanceCompFAR}, \ref{PerformanceCompFRR}, and \ref{PerformanceCompAccuracy}. By observing these figures, we can conclude two things: first the feature selection has a positive impact, as without compromising the accuracy much, we were able to get rid of more than 25\% of the features. Second, a system based on the fusion of the selected features outperforms the other variations for three out of four classifiers.\\
Is the FLF worth using? This question arises as the FLF almost doubles the number of features used to build the authentication system. The answer is, it depends. If we take only the performance (accuracy) into account, it may not be that useful. However, FLF may worth it if we consider minimal or high effort mimicry threat model as stated in section \ref{sec:ThreatModel}. Kumar et al. \cite{TreadmillAttack} suggest that authentication systems based only on acceleration are vulnerable to high effort imitation attack. They also suggest that the inclusion of other context information e.g., rotation captured through the gyroscope will add extra dimensions to the feature space. Considering the minimal or high effort mimicry threat models, we believe that fusing rotation information at the feature level may provide an extra layer of defense, in addition to improving the overall performance of the system. We do not claim that fusing rotation information will completely rule out the possibility of attack, but we believe that it will certainly make an adversary's job more difficult as fusion of rotation information adds more (31) degrees of freedom in the feature space.
\subsubsection{Score level fusion}As we used the same classifiers for both of the modalities, the classification scores were similar in nature (probabilities). The score level fusion (SLF) was one of the feasible and practical alternatives for fusing acceleration and rotation. Moreover, one of the biggest advantages of SLF is that it requires no knowledge of the underlying features extraction and classification methods \cite{SLFDescription2}\cite{SLFDescription}. We used a weighted score fusion technique in order to fuse the classification scores obtained from acceleration- and rotation-based systems built on separately on separately selected features. Let $S_a$ be the acceleration-based classification scores and $S_r$ be the rotation-based classification scores. The fused scores are computed as $S_f=(W_a\times S_a)+(W_r\times S_r)$, where $W_a$ is the weight for acceleration-based scores, $W_r$ is the weight for rotation-based scores ($W_a$ + $W_r$ = 1). In Figure \ref{SLFFAR8} and \ref{SLFFRR8}, the weights along the x-axis are the
weights assigned to the acceleration based scores whereas the weights for rotation-based are computed as $(1-W_a)$. The at the rate symbols (@) show the best combination of weights for fusion. By observing these results, we can see that SLF is not as effective as the FLF. Therefore, we conclude that feature level fusion shows more promise than the score level fusion which is not a surprise.
\subsection{Discussions}
\subsubsection{Scalability of the system}How scalable is our proposed system? Since, we tested our system on a limited number (40) of users, it is important to investigate the scalability of all three authentication systems for a larger population of users. To show that our method is scalable, we computed the error rates and accuracy starting from one user, kept on adding other users one by one, and computed the mean of error rates, and the mean of accuracies \cite{PhoneSwiping1}. The users were added in their order of participation in our data collection. We repeated the experiment until all 40 users of our database were added and tested. We can see that the mean accuracy stabilized after a certain number of users were added. For example, FLF based system achieves 95\% or higher for three (kNNEuc, LogReg, and RandFor) out of four classifiers after testing 30\% of the users (see Figures \ref{ScalabilityAcc}, \ref{ScalabilityGyro}, \ref{ScalabilityFusion}). A sharp drop of 2\% can be observed for multilayer perceptrons but soon after adding more users, the accuracy stabilized and stayed up above the 93\%. This trend suggests that all three of our systems are scalable to the desired number of users.
\begin{table}[h]
\centering
\tbl{The performance of the acceleration, rotation, and FLF-based authentication systems in intersession and interphase authentication for ONLY TWELVE common users in Phase1 and Phase2. Each sessions were separated by at least a 10 minute of interval, whereas the phases are separated by at least a three month time period. The acronyms $P_aS_b-P_cS_d$ indicate the data used for training and testing. For example, $P_1S_1-P_1S_2$ stands for Phase1 Session1 for training and Phase1 Session2 for testing. The $P_1S_1-P_1S_2$ and $P_2S_1-P_2S_2$ are for the inter session, whereas $P_1S_1-P_2S_1$ and $P_1S_1-P_2S_2$ are for the inter-phase. DAc stands for dynamic accuracy.}{
\begin{tabular}{|l|l|l|l|l|l|l|l|l|l|l|l|l|}
\hline
\multirow{4}{*}{\textbf{\begin{tabular}[c]{@{}l@{}}Class-\\ ifiers\end{tabular}}} & \multicolumn{6}{c|}{\textbf{Inter-Session}}                                                 & \multicolumn{6}{c|}{\textbf{Inter-Phase}}                                                   \\ \cline{2-13}
                                                                                  & \multicolumn{3}{c|}{\textbf{$P_1S_1-P_1S_2$}}       & \multicolumn{3}{c|}{\textbf{$P_2S_1-P_2S_2$}}       & \multicolumn{3}{c|}{\textbf{$P_1S_1-P_2S_1$}}       & \multicolumn{3}{c|}{\textbf{$P_1S_1-P_2S_2$}}       \\ \cline{2-13}
                                                                                  & \textbf{DFAR} & \textbf{DFRR} & \textbf{DAc} & \textbf{DFAR} & \textbf{DFRR} & \textbf{DAc} & \textbf{DFAR} & \textbf{DFRR} & \textbf{DAc} & \textbf{DFAR} & \textbf{DFRR} & \textbf{DAc} \\ \cline{2-13}
                                                                                  & \multicolumn{12}{c|}{\textbf{Performance of acceleration-based authentication system}}                                                                                                                   \\ \hline
\textbf{kNNEuc}                                                                   & 9.91          & 3.85          & 93.1         & 8.49          & 3.69          & 93.9         & 16.04         & 22.76         & 80.6         & 13.68         & 23.24         & 81.6         \\ \hline
\textbf{LogReg}                                                                   & 16.04         & 3.21          & 90.3         & 12.58         & 3.85          & 91.8         & 17.45         & 8.81          & 86.8         & 21.23         & 11.06         & 83.8         \\ \hline
\textbf{MulPer}                                                                   & 17.45         & 2.08          & 90.2         & 15.57         & 3.53          & 90.4         & 22.48         & 6.89          & 85.2         & 25.16         & 7.37          & 83.7         \\ \hline
\textbf{RanFor}                                                                   & 18.4          & 3.85          & 88.8         & 22.8          & 4.01          & 86.5         & 30.5          & 3.69          & 82.8         & 26.42         & 4.81          & 84.3         \\ \hline
                                                                                  & \multicolumn{12}{c|}{\textbf{Performance of rotation-based authentication system}}                                                                                                                       \\ \hline
\textbf{kNNEuc}                                                                   & 8.81          & 8.01          & 91.6         & 9.75          & 4.33          & 92.9         & 16.19         & 17.95         & 82.9         & 15.09         & 26.92         & 79.1         \\ \hline
\textbf{LogReg}                                                                   & 13.05         & 5.61          & 90.6         & 12.89         & 3.69          & 91.7         & 16.67         & 20.67         & 81.4         & 16.35         & 27.72         & 78           \\ \hline
\textbf{MulPer}                                                                   & 16.67         & 5.61          & 88.8         & 13.99         & 2.4           & 91.8         & 22.01         & 18.59         & 79.7         & 24.37         & 25.32         & 75.2         \\ \hline
\textbf{RanFor}                                                                   & 15.57         & 2.24          & 91           & 15.88         & 4.33          & 89.8         & 19.65         & 10.26         & 85           & 20.75         & 17.31         & 81           \\ \hline
                                                                                  & \multicolumn{12}{c|}{\textbf{Performance of FLF-based authentication system}}                                                                                                                            \\ \hline
\textbf{kNNEuc}                                                                   & 5.68          & 4.23          & 95           & 8.33          & 3.33          & 94.1         & 17.68         & 12.31         & 85           & 15.03         & 14.62         & 85.2         \\ \hline
\textbf{LogReg}                                                                   & 12.63         & 3.21          & 92           & 11.99         & 1.28          & 93.3         & 18.43         & 6.03          & 87.7         & 17.8          & 5.38          & 88.4         \\ \hline
\textbf{MulPer}                                                                   & 12.5          & 2.82          & 92.3         & 13.64         & 0.77          & 92.7         & 23.61         & 5.13          & 85.6         & 22.35         & 3.72          & 86.9         \\ \hline
\textbf{RanFor}                                                                   & 16.29         & 1.03          & 91.3         & 21.09         & 0.9           & 88.9         & 26.77         & 2.31          & 85.4         & 26.39         & 2.31          & 85.6         \\ \hline
\end{tabular}}
\label{InterPhaseINterSession}
\end{table}
\subsubsection{Impact of external phenomenon on the performance}
\label{sec:TemplateUpdate}In order to analyze how external phenomenon, e.g., physiological changes, shoe type, emotional state, pose and orientation, prior activity, time, etc., affect the authentication accuracy, we collected the data in two different phases separated by at least three months. We decided to use a three month period because we assume that most of the mentioned external phenomenon change significantly enough to affect the performance of the authentication system. We evaluated the performance of each system with all four classification algorithms. The performance of systems based on acceleration, rotation and fusion are presented in Table \ref{InterPhaseINterSession}. We can observe that the performance of each system in the inter-phase setting is poor compared to the one in the inter-session setting. However, the error rates are still acceptable for a continuous authentication system. The $P_2S_1-P_2S_2$  column of Table \ref{InterPhaseINterSession} presents the results of the template update. To show the effect of the template update we trained the classifier again by using the latest ($P_2S_1$) data and updated the thresholds for every user. The performance is tested by using the data collected later on i.e. in $P_2S_2$. We can see that the accuracy is almost unaltered which means if the template gets updated periodically the performance (accuracy) of the system will remain above 90 in most of the cases.
\subsubsection{Arm movements while walking as a behaviometric}We argue that arm movements while walking qualifies as a behaviometric based on these criteria: universality, distinctiveness, permanence, collectability, performance, acceptability, and circumvention \cite{JainBiometrics}. Since arm movements, while a person walks, are a universal behavior, it satisfies the universality criteria. Our experimental results support the distinctiveness and permanence criteria-- distinctiveness is supported by accuracy in authentication ($\sim$96\%) and permanence through the training and testing data used in different time frames. Because the data collection or the method of interaction with the device for authentication does not pose procedural issues, we claim that arm movements recorded through smartwatches satisfy collectability and acceptability criteria.  We do not expect drastic change in arm movements within the same context thus our claim of circumvention. Thus we posit that arm movements while walking has a potential to be a behaviometric.
\subsubsection{Application Scenarios}
The security of components (cars, mobile, tablets, computers, house, lockers, offices, schools, restaurants, airports, weapons, etc.) of the Internet of Things (IoT) has become more prevalent than ever before.  The authenticity or identity of the owner of these devices has to be continuously and unobtrusively verified, instead of the one point authentication offered by PINs, passwords, patterns, or physiological characteristics such as finger, palm, iris, face prints etc. The potential channels to verify the authenticity or identity of the owner are wearable or hand-held devices such as smartphones, smartwatches, Google glass, key fobs etc.  The proposed behaviometric in this paper can be potentially used to authenticate users for accessing the aforementioned buildings and devices.\\
For example, consider an organization like Central Intelligence Agency (CIA), where, there are various levels of security clearance that allow authorized people to access secured resources. The current existing security mechanisms verify the authenticity of a person at the time of entry. If the security service at the time of entry is compromised, the CIA does not have any other means to detect the authenticity of the individual. In this case, our proposed behaviometric can be very helpful because it provides continuous and unobtrusive verification of authorized individuals. The authorized person will have to wear a smartwatch while they walk up to the secured equipment, weapons, documents, or buildings.  The security system will automatically verify them easily, efficiently, and conveniently by using the data generated by their arm movements.  A similar concept can be seen in the movie \text{Mission:Impossible - Rogue Nation}, where a camera based gait authentication system is used to create the most secure and unbreakable system seen to date.
\section{Conclusion and Future Work}
\label{sec:Conclusion}
We conclude that arm movements while walking can be used as a viable means for authenticating users. We designed, implemented, and evaluated four continuous authentication mechanisms, first by using acceleration of arms, second rotation of arms, and third through fusion of these two at the feature level. We tested all four designs (acceleration, rotation, and fusion) by using four machine learning classification algorithms. Our experimental results suggest that, although acceleration and rotation individually are sufficiently discriminative to build an authentication system, the feature-level fusion of these modalities not only improves the overall performance of the system but it also arguably adds an extra layer of defense against potential high-effort mimicry attacks. We also conclude that the score level fusion does not improve the performance much. By performing in-depth analysis of features extracted from each acceleration and rotation, we provide the best suitable subsets of features for classification. Our empirical analysis shows that $W_{size}$ between eight to twelve seconds and $S_{Interval}$ of two to four seconds are effective in general. These settings offer more frequent and reliable authentication decisions.\\
In future, we plan to investigate the following: investigate the impact of data and score level fusion; analyze the tradeoffs between the number of features used for classification and the performance; statistical evidence of the robustness of fusion based systems against high effort mimicry attacks; examine how the proposed biometric can be used to authenticate users on smartphone paired with a smartwatch.
\section{Acknowledgements}
We would like to thank all volunteers for participating in our data collection and anonymous reviewers for their insightful feedback. All findings, conclusions and recommendations expressed in this paper are those of the authors and do not necessarily represent the views of any organization.
\bibliographystyle{acmsmall}
\bibliography{SmartWatchTissec}

\begin{thebibliography}{}

\bibitem[\protect\citeauthoryear{A and B}{A and B}{2005}]{FLF2}
{\sc A, A.~R.} {\sc and} {\sc B, R.~G.} 2005.
\newblock Feature level fusion using hand and face biometrics.
\newblock In {\em Proceedings of SPIE Conference on Biometric Technology for
  Human Identification II}. 196--204.

\bibitem[\protect\citeauthoryear{Aha and Kibler}{Aha and Kibler}{1991}]{kNNEuc}
{\sc Aha, D.} {\sc and} {\sc Kibler, D.} 1991.
\newblock Instance-based learning algorithms.
\newblock {\em Machine Learning\/}~{\em 6}, 37--66.

\bibitem[\protect\citeauthoryear{Ahmed}{Ahmed}{2008}]{BiometricsMetrics}
{\sc Ahmed, A. A. E.~S.} 2008.
\newblock Security monitoring through human computer interaction devices.
\newblock Ph.D. thesis, Victoria, B.C., Canada, Canada.
\newblock AAINR60664.

\bibitem[\protect\citeauthoryear{Breiman}{Breiman}{2001}]{RandomForest}
{\sc Breiman, L.} 2001.
\newblock Random forests.
\newblock {\em Machine Learning\/}~{\em 45,\/}~1, 5--32.

\bibitem[\protect\citeauthoryear{Buchoux and Clarke}{Buchoux and
  Clarke}{2008}]{PhoneTyping2}
{\sc Buchoux, A.} {\sc and} {\sc Clarke, N.~L.} 2008.
\newblock Deployment of keystroke analysis on a smartphone.
\newblock In {\em Australian Information Security Management Conference}. 48.

\bibitem[\protect\citeauthoryear{Charara}{Charara}{2015}]{SmartwatchPayments}
{\sc Charara, S.} 2015.
\newblock Samsung's bet on biometrics for smartwatch payments.
\newblock
  \url{http://www.ibtimes.co.uk/apple-iphone-5s-touch-id-fingerprint-scanner-508196}.
\newblock [Online; Last accessed in 18 July, 2015].

\bibitem[\protect\citeauthoryear{Charlton}{Charlton}{2013}]{FingerPrintScanner}
{\sc Charlton, A.} 2013.
\newblock iphone 5s fingerprint security bypassed by german computer club.
\newblock
  \url{http://www.ibtimes.co.uk/apple-iphone-5s-touch-id-fingerprint-scanner-508196}.
\newblock [Online; Last accessed 18 July, 2015].

\bibitem[\protect\citeauthoryear{Curtis}{Curtis}{2014}]{ReplaceSmartphones}
{\sc Curtis}. 2014.
\newblock Why smartwatches will replace smartphones: biometrics and
  convenience.
\newblock
  \url{http://www.ibtimes.co.uk/apple-iphone-5s-touch-id-fingerprint-scanner-508196}.
\newblock [Online; Last accessed in 18 July, 2015].

\bibitem[\protect\citeauthoryear{Dass, N, and Jain}{Dass
  et~al\mbox{.}}{2005}]{SLFDescription2}
{\sc Dass, S.~C.}, {\sc N, K.}, {\sc and} {\sc Jain, A.~K.} 2005.
\newblock A principled approach to score level fusion in multimodal biometric
  systems.
\newblock In {\em Proceedings of AVBPA}. 1049--1058.

\bibitem[\protect\citeauthoryear{Derawi, Bours, and Holien}{Derawi
  et~al\mbox{.}}{2010}]{PhoneGait2}
{\sc Derawi, M.}, {\sc Bours, P.}, {\sc and} {\sc Holien, K.} 2010.
\newblock Improved cycle detection for accelerometer based gait authentication.
\newblock In {\em IIH-MSP, 2010 Sixth International Conference on}. 312--317.

\bibitem[\protect\citeauthoryear{Derawi, Nickel, Bours, and Busch}{Derawi
  et~al\mbox{.}}{2010}]{AuthenticationForPhones}
{\sc Derawi, M.~O.}, {\sc Nickel, C.}, {\sc Bours, P.}, {\sc and} {\sc Busch,
  C.} 2010.
\newblock Unobtrusive user-authentication on mobile phones using biometric gait
  recognition.
\newblock [Online; Last accessed 04 August, 2015].

\bibitem[\protect\citeauthoryear{Dernbach, Das, Krishnan, Thomas, and
  Cook}{Dernbach et~al\mbox{.}}{2012}]{ActivityRecognition}
{\sc Dernbach, S.}, {\sc Das, B.}, {\sc Krishnan, N.~C.}, {\sc Thomas, B.},
  {\sc and} {\sc Cook, D.} 2012.
\newblock Simple and complex activity recognition through smart phones.
\newblock In {\em Intelligent Environments (IE), 2012 8th International
  Conference on}. 214--221.

\bibitem[\protect\citeauthoryear{Fraccaro, Coyle, Doyle, and
  O'Sullivan}{Fraccaro et~al\mbox{.}}{2014}]{ETelemed}
{\sc Fraccaro, P.}, {\sc Coyle, L.}, {\sc Doyle, J.}, {\sc and} {\sc
  O'Sullivan, D.} 2014.
\newblock Real-world gyroscope-based gait event detection and gait feature
  extraction.

\bibitem[\protect\citeauthoryear{Frank, Biedert, Ma, Martinovic, and
  Song}{Frank et~al\mbox{.}}{2013}]{PhoneSwiping1}
{\sc Frank, M.}, {\sc Biedert, R.}, {\sc Ma, E.}, {\sc Martinovic, I.}, {\sc
  and} {\sc Song, D.} 2013.
\newblock Touchalytics: On the applicability of touchscreen input as a
  behavioral biometric for continuous authentication.
\newblock {\em Information Forensics and Security, IEEE Transactions on\/}~{\em
  8,\/}~1, 136--148.

\bibitem[\protect\citeauthoryear{Gafurov, Snekkenes, and Bours}{Gafurov
  et~al\mbox{.}}{2007}]{Gafurov}
{\sc Gafurov, D.}, {\sc Snekkenes, E.}, {\sc and} {\sc Bours, P.} 2007.
\newblock Spoof attacks on gait authentication system.
\newblock {\em Information Forensics and Security, IEEE Transactions on\/}~{\em
  2,\/}~3, 491--502.

\bibitem[\protect\citeauthoryear{Greyb}{Greyb}{2015}]{AuthenticatePhoneWithWatch}
{\sc Greyb}. 2015.
\newblock Samsung’s smartwatch: Your heartbeats will unlock your next
  smartphone.
\newblock
  \url{http://www.whatafuture.com/2015/03/17/heartbeat-authentication-in-samsung-smartwatch/#sthash.gWOBeptQ.AbPqCxwm.dpbs}.
\newblock [Online; Last accessed in 18 July, 2015].

\bibitem[\protect\citeauthoryear{Guiry, van~de Ven, and Nelson}{Guiry
  et~al\mbox{.}}{2014}]{FusionOfMultipleSensorsAccGyro}
{\sc Guiry, J.~J.}, {\sc van~de Ven, P.}, {\sc and} {\sc Nelson, J.} 2014.
\newblock Multi-sensor fusion for enhanced contextual awareness of everyday
  activities with ubiquitous devices.
\newblock {\em Sensors\/}~{\em 14,\/}~3, 5687.

\bibitem[\protect\citeauthoryear{Hall and Holmes}{Hall and
  Holmes}{2003}]{BenchmarkingAttributeSelectors}
{\sc Hall, M.} {\sc and} {\sc Holmes, G.} 2003.
\newblock Benchmarking attribute selection techniques for discrete class data
  mining.
\newblock {\em Knowledge and Data Engineering, IEEE Transactions on\/}~{\em
  15,\/}~6, 1437--1447.

\bibitem[\protect\citeauthoryear{Hall}{Hall}{1998}]{CfsEvaluator}
{\sc Hall, M.~A.} 1998.
\newblock Correlation-based feature subset selection for machine learning.
\newblock Ph.D. thesis, University of Waikato, Hamilton, New Zealand.

\bibitem[\protect\citeauthoryear{Hassoun}{Hassoun}{1995}]{MultilayerPerceptrons}
{\sc Hassoun, M.~H.} 1995.
\newblock {\em Fundamentals of Artificial Neural Networks\/} 1st Ed.
\newblock MIT Press, Cambridge, MA, USA.

\bibitem[\protect\citeauthoryear{He, Horng, Fan, Run, Chen, Lai, Khan, and
  Sentosa}{He et~al\mbox{.}}{2010}]{SLFDescription}
{\sc He, M.}, {\sc Horng, S.-J.}, {\sc Fan, P.}, {\sc Run, R.-S.}, {\sc Chen,
  R.-J.}, {\sc Lai, J.-L.}, {\sc Khan, M.~K.}, {\sc and} {\sc Sentosa, K.~O.}
  2010.
\newblock Performance evaluation of score level fusion in multimodal biometric
  systems.
\newblock {\em Pattern Recogn.\/}~{\em 43,\/}~5, 1789--1800.

\bibitem[\protect\citeauthoryear{IoT}{IoT}{2016}]{WikipediaIoT}
{\sc IoT}. 2016.
\newblock Internet of things.
\newblock \url{https://en.wikipedia.org/wiki/Internet_of_Things}.
\newblock [Online; Last accessed 9 February, 2016].

\bibitem[\protect\citeauthoryear{Jain and Ross}{Jain and
  Ross}{2002}]{UserSpecificJain}
{\sc Jain, A.} {\sc and} {\sc Ross, A.} 2002.
\newblock Learning user-specific parameters in a multibiometric system.
\newblock In {\em Image Processing. 2002. Proceedings. 2002 International
  Conference on}. Vol.~1. I--57--I--60 vol.1.

\bibitem[\protect\citeauthoryear{Jain, Ross, and Prabhakar}{Jain
  et~al\mbox{.}}{2004}]{JainBiometrics}
{\sc Jain, A.~K.}, {\sc Ross, A.}, {\sc and} {\sc Prabhakar, S.} 2004.
\newblock An introduction to biometric recognition.
\newblock {\em IEEE Trans. on Circuits and Systems for Video Technology\/}~{\em
  14}, 4--20.

\bibitem[\protect\citeauthoryear{Johnston and Weiss}{Johnston and
  Weiss}{2015}]{FordhamSmartWatch}
{\sc Johnston, A.} {\sc and} {\sc Weiss, G.} 2015.
\newblock Smartwatch-based biometric gait recognition.
\newblock In {\em Biometrics Theory, Applications and Systems (BTAS), 2015 IEEE
  7th International Conference on}. 1--6.

\bibitem[\protect\citeauthoryear{Junshuang~Yang}{Junshuang~Yang}{2015}]{HMMSmartWatch}
{\sc Junshuang~Yang, Yanyan~Li, M.~X.} 2015.
\newblock Motionauth: Motion-based authentication for wrist worn smart devices.
\newblock In {\em Workshop on Sensing Systems and Applications Using Wrist Worn
  Smart Devices}. 1049--1058.

\bibitem[\protect\citeauthoryear{Kumar, Phoha, and Jain}{Kumar
  et~al\mbox{.}}{2015}]{TreadmillAttack}
{\sc Kumar, R.}, {\sc Phoha, V.}, {\sc and} {\sc Jain, A.} 2015.
\newblock Treadmill assisted imitation attack on gait-based authentication
  systems.
\newblock In {\em Biometrics: Theory, Applications and Systems (BTAS), 2015
  IEEE Seventh International Conference on}. 1--7.

\bibitem[\protect\citeauthoryear{Kwapisz, Weiss, and Moore}{Kwapisz
  et~al\mbox{.}}{2010}]{PhoneGait1}
{\sc Kwapisz, J.}, {\sc Weiss, G.}, {\sc and} {\sc Moore, S.} 2010.
\newblock Cell phone-based biometric identification.
\newblock In {\em Biometrics: Theory Applications and Systems (BTAS), 2010
  Fourth IEEE International Conference on}. 1--7.

\bibitem[\protect\citeauthoryear{Kwapisz, Weiss, and Moore}{Kwapisz
  et~al\mbox{.}}{2011}]{KwapiszActivity}
{\sc Kwapisz, J.~R.}, {\sc Weiss, G.~M.}, {\sc and} {\sc Moore, S.~A.} 2011.
\newblock Activity recognition using cell phone accelerometers.
\newblock {\em SIGKDD Explor. Newsl.\/}~{\em 12,\/}~2, 74--82.

\bibitem[\protect\citeauthoryear{MATLAB}{MATLAB}{013a}]{MATLAB2013a}
{\sc MATLAB}. 2013a.
\newblock {\em version 8.1.0.604 (R2013a)}.
\newblock The MathWorks Inc., Natick, Massachusetts.

\bibitem[\protect\citeauthoryear{Monroe}{Monroe}{}]{BiometricsMetricsReport}
{\sc Monroe, D.}
\newblock Biometrics metric report.
\newblock \url{http://www.usma.edu/ietd/docs/BiometricsMetricsReport.pdf}.
\newblock [Online; Last accessed 27 November, 2015].

\bibitem[\protect\citeauthoryear{Nagar, Nandakumar, and Jain}{Nagar
  et~al\mbox{.}}{2012}]{FLF1}
{\sc Nagar, A.}, {\sc Nandakumar, K.}, {\sc and} {\sc Jain, A.} 2012.
\newblock Multibiometric cryptosystems based on feature-level fusion.
\newblock {\em Information Forensics and Security, IEEE Transactions on\/}~{\em
  7,\/}~1, 255--268.

\bibitem[\protect\citeauthoryear{Nickel, Wirtl, and Busch}{Nickel
  et~al\mbox{.}}{2012}]{kNNForGait}
{\sc Nickel, C.}, {\sc Wirtl, T.}, {\sc and} {\sc Busch, C.} 2012.
\newblock Authentication of smartphone users based on the way they walk using
  k-nn algorithm.
\newblock In {\em Intelligent Information Hiding and Multimedia Signal
  Processing (IIH-MSP), 2012 Eighth International Conference on}. 16--20.

\bibitem[\protect\citeauthoryear{Niinuma, Park, and Jain}{Niinuma
  et~al\mbox{.}}{2010}]{ContinuousAuthJain}
{\sc Niinuma, K.}, {\sc Park, U.}, {\sc and} {\sc Jain, A.} 2010.
\newblock Soft biometric traits for continuous user authentication.
\newblock {\em Information Forensics and Security, IEEE Transactions on\/}~{\em
  5,\/}~4, 771--780.

\bibitem[\protect\citeauthoryear{Porzi, Messelodi, Modena, and Ricci}{Porzi
  et~al\mbox{.}}{2013}]{SmartWatchGestureRecognition}
{\sc Porzi, L.}, {\sc Messelodi, S.}, {\sc Modena, C.~M.}, {\sc and} {\sc
  Ricci, E.} 2013.
\newblock A smart watch-based gesture recognition system for assisting people
  with visual impairments.
\newblock In {\em Proceedings of the 3rd ACM International Workshop on
  Interactive Multimedia on Mobile \&\#38; Portable Devices}. IMMPD '13. ACM,
  New York, NY, USA, 19--24.

\bibitem[\protect\citeauthoryear{Primo, Phoha, Kumar, and Serwadda}{Primo
  et~al\mbox{.}}{2014}]{ABENA}
{\sc Primo, A.}, {\sc Phoha, V.~V.}, {\sc Kumar, R.}, {\sc and} {\sc Serwadda,
  A.} 2014.
\newblock Context-aware active authentication using smartphone accelerometer
  measurements.
\newblock {\em The IEEE Conference on Computer Vision and Pattern Recognition
  (CVPR) Workshops\/}.

\bibitem[\protect\citeauthoryear{Ratha, Connell, and Bolle}{Ratha
  et~al\mbox{.}}{2001}]{AttackSources}
{\sc Ratha, N.~K.}, {\sc Connell, J.~H.}, {\sc and} {\sc Bolle, R.~M.} 2001.
\newblock An analysis of minutiae matching strength.
\newblock In {\em Proceedings of the Third International Conference on Audio-
  and Video-Based Biometric Person Authentication}. AVBPA '01. Springer-Verlag,
  London, UK, UK, 223--228.

\bibitem[\protect\citeauthoryear{Rattani, Poh, and Ross}{Rattani
  et~al\mbox{.}}{2012}]{UserSpecificSpoofProof}
{\sc Rattani, A.}, {\sc Poh, N.}, {\sc and} {\sc Ross, A.} 2012.
\newblock Analysis of user-specific score characteristics for spoof biometric
  attacks.
\newblock In {\em Computer Vision and Pattern Recognition Workshops (CVPRW),
  2012 IEEE Computer Society Conference on}. 124--129.

\bibitem[\protect\citeauthoryear{Sarkisyan, Debbiny, and Nahapetian}{Sarkisyan
  et~al\mbox{.}}{2015}]{WristSnoop}
{\sc Sarkisyan, A.}, {\sc Debbiny, R.}, {\sc and} {\sc Nahapetian, A.} 2015.
\newblock Wristsnoop: Smartphone pins prediction using smartwatch motion
  sensors.
\newblock In {\em Information Forensics and Security (WIFS), 2015 IEEE
  International Workshop on}. 1--6.

\bibitem[\protect\citeauthoryear{Serwadda, Phoha, and Wang}{Serwadda
  et~al\mbox{.}}{2013}]{PhoneSwiping2}
{\sc Serwadda, A.}, {\sc Phoha, V.}, {\sc and} {\sc Wang, Z.} 2013.
\newblock Which verifiers work?: A benchmark evaluation of touch-based
  authentication algorithms.
\newblock In {\em Biometrics: Theory, Applications and Systems (BTAS), 2013
  IEEE Sixth International Conference on}. 1--8.

\bibitem[\protect\citeauthoryear{Serwadda and Phoha}{Serwadda and
  Phoha}{2013}]{SerwaddaRobotics}
{\sc Serwadda, A.} {\sc and} {\sc Phoha, V.~V.} 2013.
\newblock When kids' toys breach mobile phone security.
\newblock In {\em Proceedings of the 2013 ACM SIGSAC Conference on Computer
  \&\#38; Communications Security}. CCS '13. ACM, New York, NY, USA, 599--610.

\bibitem[\protect\citeauthoryear{Shukla, Kumar, Serwadda, and Phoha}{Shukla
  et~al\mbox{.}}{2014}]{ShuklaAttack}
{\sc Shukla, D.}, {\sc Kumar, R.}, {\sc Serwadda, A.}, {\sc and} {\sc Phoha,
  V.~V.} 2014.
\newblock Beware, your hands reveal your secrets!
\newblock In {\em Proceedings of the 2014 ACM SIGSAC Conference on Computer and
  Communications Security}. CCS '14. ACM, New York, NY, USA, 904--917.

\bibitem[\protect\citeauthoryear{Traore, Traore, and Ahmed}{Traore
  et~al\mbox{.}}{2011}]{ContinuousAuthentication}
{\sc Traore, I.}, {\sc Traore, I.}, {\sc and} {\sc Ahmed, A. A.~E.} 2011.
\newblock {\em Continuous Authentication Using Biometrics: Data, Models, and
  Metrics\/} 1st Ed.
\newblock IGI Global, Hershey, PA, USA.

\bibitem[\protect\citeauthoryear{Witten and Frank}{Witten and
  Frank}{2005}]{LogisticRegression}
{\sc Witten, I.~H.} {\sc and} {\sc Frank, E.} 2005.
\newblock {\em Data Mining: Practical Machine Learning Tools and Techniques,
  Second Edition (Morgan Kaufmann Series in Data Management Systems)}.
\newblock Morgan Kaufmann Publishers Inc., San Francisco, CA, USA.

\bibitem[\protect\citeauthoryear{Yan, Subbaraju, Chakraborty, Misra, and
  Aberer}{Yan et~al\mbox{.}}{2012}]{EnergyEfficient}
{\sc Yan, Z.}, {\sc Subbaraju, V.}, {\sc Chakraborty, D.}, {\sc Misra, A.},
  {\sc and} {\sc Aberer, K.} 2012.
\newblock Energy-efficient continuous activity recognition on mobile phones: An
  activity-adaptive approach.
\newblock In {\em Wearable Computers (ISWC), 2012 16th International Symposium
  on}. 17--24.

\bibitem[\protect\citeauthoryear{Zhong and Deng}{Zhong and
  Deng}{2014}]{GDIBasedGBAS}
{\sc Zhong, Y.} {\sc and} {\sc Deng, Y.} 2014.
\newblock Sensor orientation invariant mobile gait biometrics.
\newblock In {\em Biometrics (IJCB), 2014 IEEE International Joint Conference
  on}. 1--8.

\end{thebibliography}
\end{document}